\documentclass[10pt,twocolumn,letterpaper]{article}

\usepackage[pagenumbers]{cvpr}      

\usepackage{cuted}

\usepackage{currfile}

\usepackage{caption}

\definecolor{cvprblue}{rgb}{0.21,0.49,0.74}
\usepackage[pagebackref,breaklinks,colorlinks,allcolors=cvprblue]{hyperref}

\title{BioHuman: Learning Biomechanical Human Representations from Video}

\author{First Author\\
Institution1\\
Institution1 address\\
{\tt\small firstauthor@i1.org}
\and
Second Author\\
Institution2\\
First line of institution2 address\\
{\tt\small secondauthor@i2.org}
}

\author{Yujun Huo$^{1,*}$ \quad He Zhang$^{2,\dagger}$ \quad Chentao Song$^{2}$ \quad Honglin Song$^{2}$\\ Ming Yan$^{2}$ \quad Jiachen Zhao$^{2}$ \quad Zongyu Zuo$^{1}$ \quad Tao Yu$^{2}$\\ $^{1}$Beihang University, China \quad $^{2}$Tsinghua University, China\\ {\small $^{\dagger}$Corresponding authors.} {\small hector\_z@126.com.}} \begin{document}

\twocolumn[{%
\renewcommand\twocolumn[1][!htb]{#1}%
\maketitle
\vspace{-9mm}
\begin{center}
    \centering
    \includegraphics[width=1.0\linewidth]{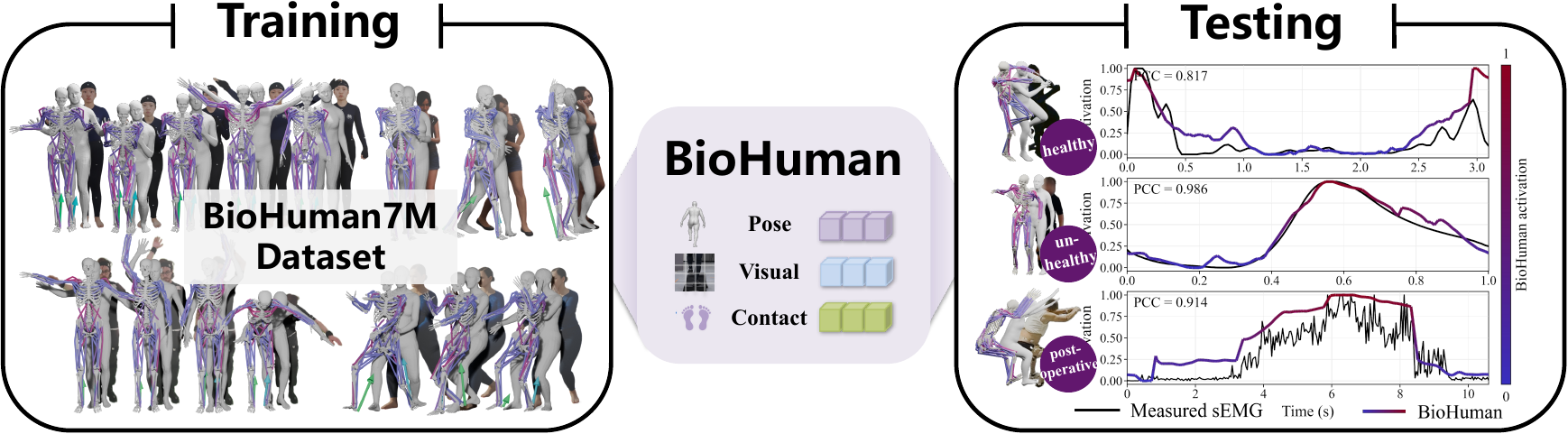}
    \vspace{-6mm}
    \captionof{figure}{\textbf{BioHuman7M} is a \textbf{Bio}mechanical \textbf{Human} dataset
    with \textbf{7.46M} paired frames, aligning real-world visual motion with
    physics-based musculoskeletal states.
    It contains representative action sequences from four datasets, including
    (a) MotionPRO, (b) EMDB, (c) Human3.6M, and (d) BEDLAM,
    with synchronized real images, SMPL bodies, and OpenSim musculoskeletal
    models annotated with ground reaction forces and muscle activations.
    Muscle color encodes activation, i.e., the degree of muscle exertion,
    with redder muscles indicating stronger activation.
    Trained on BioHuman7M, our \textbf{BioHuman} model estimates the activations of 112 muscles end-to-end from monocular video, demonstrating strong performance on healthy subjects, individuals with cerebral palsy, and postoperative patients.}

    \label{fig:teaser}
    \vspace{-1mm}
\end{center}%
}]

\begin{abstract}
Muscle activations provide insights into the muscular mechanisms underlying human movement, yet traditional motion analysis remains focused on body geometry and kinematics. Existing muscle activation estimation methods rely on intermediate motion representations that limit access to visual cues at the image-level. This challenge is further compounded by the lack of large-scale datasets pairing visual observations with muscle activation. To bridge vision and biomechanics, we develop BioSim to generate musculoskeletal annotations from SMPL motion and construct BioHuman7M, a 7.46-million-frame dataset with aligned visual and biomechanical information. Building on this dataset, we propose BioHuman, an end-to-end video-to-muscle framework that jointly integrates kinematic representations, image-level visual evidence, and contact context to predict full-body muscle activations. A temporal transformer processes the corresponding pose, visual, and contact tokens to directly estimate 112-dimensional muscle activations. Experiments demonstrate state-of-the-art performance on simulation-based benchmarks and strong agreement with measured surface electromyography (sEMG) signals from rehabilitation and clinical datasets. These results highlight the potential of the proposed method for gait rehabilitation and functional movement assessment.
\end{abstract}

\section{Introduction}
\label{sec:intro}
Muscle activations reveal the muscular mechanisms underlying human movement, providing important insights into movement impairments and rehabilitation outcomes. While video-based motion capture~\cite{goel2023hmrv2,kocabas2020vibe,shin2024wham,wang2024tram,shen2024gvhmr,wang2025prompthmr,zhang2024proxycap} has advanced human geometry and kinematic reconstruction using parametric models such as SMPL~\cite{loper2015smpl}, these representations describe how the body moves rather than how muscles generate movement. Estimating muscle activations from video would extend motion capture towards characterizing the muscular processes underlying movement. 
However, this task remains challenging because muscles are not directly observable from visual inputs, and their activation cannot be reliably inferred from kinematics alone.

A key challenge is the lack of large-scale datasets pairing visual observations with full-body muscle activity annotations. Existing human datasets~\cite{ionescu2013h36m,von20183dpw,black2023bedlam,kaufmann2023emdb,ren2025motionpro} primarily provide images and body motion, without dense full-body muscle activity annotations. Muscle activity data requires EMG systems, which typically capture activity from only a subset of muscles~\cite{schreiber2019multimodal,fernandez2026lgnk,dabrowski2026humaps}. Consequently, existing biomechanics-aware methods remain limited in scale, motion coverage, or anatomical scope~\cite{chiquier2023MiA,schneider2024mint}.

To address this data gap, we introduce BioHuman7M, a large-scale dataset comprising 7.46 million frames, constructed by augmenting established motion datasets, including Human3.6M~\cite{ionescu2013h36m}, BEDLAM~\cite{black2023bedlam}, EMDB~\cite{kaufmann2023emdb}, and MotionPRO~\cite{ren2025motionpro}, with musculoskeletal annotations. We develop BioSim, a pipeline that retargets SMPL motion to an OpenSim musculoskeletal model, estimates ground-reaction forces, and derives 112-dimensional muscle activation labels. This process integrates visual observations, body motion, contact-related dynamics, and muscle activations into a temporally aligned dataset. To assess the physiological plausibility of the derived activations, we evaluate BioSim on datasets with synchronized sEMG recordings by comparing estimated activation patterns with measured signals from corresponding muscles.

While BioHuman7M provides large-scale paired supervision, leveraging visual information for muscle activation estimation remains challenging. A common strategy for video-based muscle activation estimation is to adopt a two-stage pipeline~\cite{chiquier2023MiA,schneider2024mint}, first recovering human pose and motion from video and then inferring muscle activations from the estimated kinematics. However, this sequential formulation introduces an information bottleneck: compressing visual observations into intermediate motion representations inevitably discards some image-level evidence that may be relevant to muscle activation estimation. Consequently, the subsequent muscle estimation stage relies primarily on kinematic information, without direct access to the full visual context of the original video. 

Our insights are:  (1) human motion is closely associated with underlying muscle activations, making kinematic representations informative for muscle activation estimation; (2) visual observations preserve image-level evidence that may be lost in reconstructed motion representations; (3) contact states provide additional physical context about environmental support and load transfer. These insights motivate a unified framework that integrates kinematic, visual, and contact information for muscle activation estimation.

We propose BioHuman, an end-to-end video-to-muscle framework that leverages visual, kinematic, and contact information for muscle activation estimation. Specifically, BioHuman introduces three complementary token types: pose tokens encode body articulation and temporal derivatives, visual tokens preserve image-conditioned features from a pretrained human mesh recovery (HMR) front-end, and contact tokens represent the physical context of body support and environmental interaction. A temporal transformer jointly processes these token streams and estimates muscle activations. By retaining visual and contact evidence alongside kinematic representations, BioHuman enables information from different sources to contribute to muscle activation estimation.

We conduct extensive experiments to evaluate BioHuman on both simulation-based benchmarks and real-world EMG datasets. BioHuman achieves state-of-the-art (SoTA) performance on simulated data and demonstrates strong agreement with measured EMG signals, outperforming existing approaches. Notably, evaluations on EMG datasets from rehabilitation and clinical settings demonstrate its potential for gait rehabilitation and functional movement assessment through video-based muscle activation analysis.

Our contributions are summarized as follows:
\begin{itemize}
  \item We introduce BioHuman7M, a large-scale dataset comprising 7.46 million frames, which augments existing motion datasets with temporally aligned simulation-derived full-body muscle activations and contact-related biomechanical annotations.
  \item We propose BioHuman, an end-to-end monocular video-to-muscle framework that integrates visual, pose, and contact information through a temporal transformer to directly estimate 112-dimensional muscle activations.
  \item We demonstrate state-of-the-art performance on simulation-based benchmarks and validate predicted muscle activation patterns against measured surface electromyography from rehabilitation and clinical datasets. These results highlight the potential of video-based muscle activation estimation for gait rehabilitation and functional movement assessment.
\end{itemize}

\section{Related work}\label{sec:relatedwork}
\subsection{Anatomically and biomechanically informed human modeling}
Existing motion capture systems primarily recover geometric and kinematic representations of human motion from visual observations. Parametric body models such as SMPL~\cite{loper2015smpl} have substantially advanced monocular human pose and shape estimation~\cite{kanazawa2018hmr,zhang2021pymaf,zhang2024proxycap}, but provide limited insight into the underlying physiological mechanisms.

To improve anatomical and biomechanical plausibility, subsequent works introduce anatomical constraints and physical modeling into motion capture~\cite{bittner2022towards,lin20243d,pagnon2022pose2sim}. For example, MANIKIN~\cite{jiang2024manikin} incorporates joint angle limits to regularize reconstructed motions and prevent anatomically implausible poses.
In parallel, anatomically informed models such as OSSO~\cite{keller2022osso}, BOSS~\cite{shetty2023boss}, SKEL~\cite{keller2023skel}, and SOMA~\cite{alvarado2026soma} introduce explicit skeletal structures and anatomical representations. Building on these developments, HSMR~\cite{xia2025hsmr} further incorporates anatomical constraints into human motion reconstruction.

Despite these developments, existing approaches improve the anatomical and kinematic plausibility of reconstructed motion, while internal muscular states, particularly full-body muscle activations, remain largely unexplored.

\begin{table}[t]
\centering
\caption{Comparison between BioHuman7M and representative related datasets (GRF: ground reaction force; M.A.: muscle activations).}
\label{tab:datasets}
\small
\setlength{\tabcolsep}{3pt}
\begin{tabular}{lccccc}
  \toprule
  Dataset & Frames & Image & Motion & GRF & M.A. \\
  \midrule
  Human3.6M          & 3.6M        & $\checkmark$ & $\checkmark$ & $\times$ & $\times$ \\
  AMASS              & $\sim$10M  & $\times$  & $\checkmark$ & $\times$ & $\times$ \\
  GroundLink         & 1.59M       & $\times$  & $\checkmark$ & $\checkmark$ & $\times$ \\
  Van H. et al.      & 1.3M        & $\times$  & $\checkmark$ & $\checkmark$ & $\checkmark$ \\
  FLEX               & 1.8M        & $\checkmark$ & $\checkmark$ & $\times$ & $\times$ \\
  MiA                & 0.45M       & $\checkmark$ & $\times$ & $\times$ & $\times$ \\
  MinT               & 1.8M        & $\times$  & $\checkmark$ & $\checkmark$ & $\checkmark$ \\
  \midrule
  BioHuman7M (ours) & 7.46M         & $\checkmark$ & $\checkmark$ & $\checkmark$ & $\checkmark$ \\
  \bottomrule
\end{tabular}
\end{table}

\subsection{Biomechanics-aware dataset}
As summarized in Tab.~\ref{tab:datasets}, existing human datasets provide varying degrees of visual, kinematic, and biomechanical annotations. Motion capture datasets, such as Human3.6M~\cite{ionescu2013h36m}, 3DPW~\cite{von20183dpw}, EMDB~\cite{kaufmann2023emdb}, and BEDLAM~\cite{black2023bedlam}, provide visual observations and motion annotations but lack biomechanical information. Among datasets with biomechanical annotations, GroundLink~\cite{han2023groundlink} augments SMPL sequences with ground-reaction forces, while the dataset introduced by Van Hooren et al.~\cite{van2024dataset} provides motion and biomechanical measurements without paired visual observations. FLEX~\cite{yin2025flex} and MiA~\cite{chiquier2023MiA} pair video with measured sEMG, but provide surface electrical signals from localized body regions rather than full-body muscle activations. In contrast, MinT~\cite{schneider2024mint} uses musculoskeletal simulation to annotate AMASS~\cite{mahmood2019amass} motion sequences with muscle activations, but lacks paired visual data.

Separately, datasets such as CAMS-Knee~\cite{taylor2017camsknee}, LIVE-GaitNeuroKids~\cite {fernandez2026lgnk}, and HUMAPS-4D~\cite{dabrowski2026humaps} provide measured sEMG alongside human movement data, offering valuable insights into muscle electrical activity and neuromuscular coordination. However, sEMG signals are related to muscle activations but do not directly quantify normalized muscle activation, and measurements typically cover only a subset of muscles. 

Existing datasets still lack large-scale, temporally aligned visual observations, motion, ground-reaction forces (GRFs), and dense full-body muscle activation annotations.

\subsection{Biomechanical estimation}
Recent studies have extended human motion analysis towards biomechanical estimation. AddBiomechanics~\cite{werling2023addbiomechanics} automates conventional musculoskeletal analysis from motion capture data, while MinT~\cite{schneider2024mint} learns muscle activations from motion sequences. More recently, OpenCap~\cite{uhlrich2023opencap} and OpenCap Monocular~\cite{gilon2026opencapM} extend biomechanical analysis to video input through OpenSim~\cite{delp2007opensim}-based musculoskeletal modeling and optimization, which introduces substantial computational overhead. OpenCapBench~\cite{gozlan2025opencapbench} and VID~\cite{chen2025vid} further provide benchmarks for evaluating biomechanical estimation from visual observations. However, existing approaches generally rely on intermediate kinematic representations and sequential biomechanical processing.

In contrast, BioHuman directly estimates full-body muscle activations from monocular video by jointly leveraging kinematic, visual, and physical interaction cues, without requiring computationally intensive musculoskeletal optimization during inference.

\section{BioHuman7M}\label{sec:BioHuman7M}

\begin{figure*}
  \includegraphics[width=\textwidth]{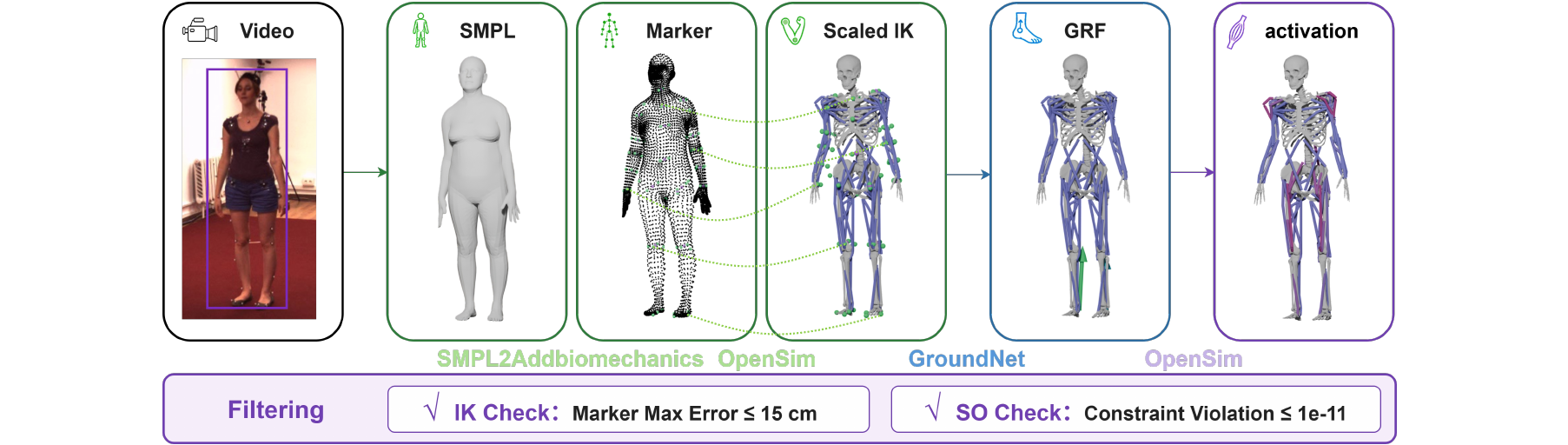}
  \caption{Pipeline of BioSim. Starting from SMPL motion sequences, we extract marker trajectories, scale the musculoskeletal model, and perform inverse kinematics to obtain OpenSim-compatible kinematics. Ground-reaction forces are estimated with GroundNet, followed by inverse dynamics and Static Optimization to compute muscle activations. The generated data are further filtered based on biomechanical errors.}
  \  \label{fig:3_dataset}
\end{figure*}

Existing human motion datasets~\cite{ionescu2013h36m,von20183dpw,kaufmann2023emdb,black2023bedlam} provide visual observations and body motion but lack large-scale, full-body biomechanical annotations. To bridge this gap, we develop BioSim (\textbf{Bio}mechanics \textbf{Sim}ulation), a pipeline that converts SMPL motion into OpenSim-compatible kinematics, estimates ground-reaction forces, and derives full-body muscle activations through musculoskeletal simulation. Applying BioSim to existing motion datasets, we construct BioHuman7M, a large-scale dataset comprising 7.46 million temporally aligned frames with visual observations, body motion, contact-related dynamics, and 112-dimensional muscle activation annotations. Figure~\ref{fig:3_dataset} illustrates the BioSim pipeline.

\subsection{SMPL to OpenSim kinematics retargeting}
To enable musculoskeletal simulation from SMPL sequences, we develop ULBS-112, a full-body musculoskeletal model for OpenSim 4.0 comprising 112 muscles, designed to balance anatomical coverage and computational efficiency at scale~\cite{zuo2024msh700}.

We establish anatomical correspondence between SMPL and ULBS-112 by defining bony landmarks on the musculoskeletal model and corresponding virtual markers on SMPL. Using SMPL2AddBiomechanics~\cite{keller2023skel}, we extract marker trajectories from SMPL sequences. OpenSim~\cite{delp2007opensim} estimates subject-specific anthropometric properties, including skeletal dimensions and muscle geometry, from the marker trajectories, body mass and height. These estimates are then used to scale the template ULBS-112 model.

Finally, we perform inverse kinematics (IK) in OpenSim on the scaled model to estimate joint-angle trajectories that best fit the SMPL-derived virtual marker trajectories. The resulting OpenSim-compatible kinematics serve as the basis for GRF estimation and muscle activation computation.

\subsection{Kinematics to GRF}\label{subsec:groundnet}
GRFs provide the external contact forces required for inverse dynamics, enabling the computation of joint moments and subsequent muscle activation estimation. However, GRFs are unavailable in SMPL motion sequences. To address this, we develop GroundNet, a temporal network that estimates bilateral GRFs and foot-contact states from SMPL motion sequences. Specifically, GroundNet projects kinematic features into a latent space and employs a Transformer encoder with positional embeddings to capture temporal dependencies. Two task-specific heads predict bilateral three-dimensional GRFs and foot-contact probabilities. 
GroundNet is trained on the GroundLink~\cite{han2023groundlink} dataset, which provides synchronized human motion, bilateral GRFs, and foot-contact annotations across diverse activities. Body-weight-normalized GRFs and foot-contact labels serve as supervision, respectively.

During inference, predictions are aggregated and scaled by the estimated body mass. Contact probabilities suppress forces during non-support phases, while force application points are derived from the bilateral heel positions computed by forward kinematics of the subject-specific OpenSim model. The resulting external loads are temporally aligned with the IK trajectories and expressed in the coordinate system of our full-body OpenSim model.

\subsection{Muscle activation simulation}
Given the musculoskeletal kinematics \(q(t)\) and estimated ground-reaction forces \(F_{\mathrm{GRF}}(t)\), we first compute the net joint moments through inverse dynamics (ID):
\begin{equation}
\tau(t) = \mathrm{ID}\big(q(t), F_{\mathrm{GRF}}(t)\big),
\label{eq:ID}
\end{equation}

We then estimate muscle activations using static optimization (SO). Since the musculoskeletal system is redundant, multiple muscle recruitment patterns can generate the same joint moments. SO resolves this redundancy by minimizing muscle activation effort at each frame:
\begin{equation}
\begin{aligned}
\min_{\{a_i(t)\}} \quad 
& \sum_{i=1}^{N_m} a_i^p(t) \\
\mathrm{s.t.} \quad
& \tau(t) =
\sum_{i=1}^{N_m} r_i(q(t))
F_i\big(a_i(t), q(t), \dot{q}(t)\big)
+ \tau_{\mathrm{res}}(t), \\
& 0 \leq a_i(t) \leq 1,
\quad i = 1,\ldots,N_m.
\end{aligned}
\label{eq:static_optimization}
\end{equation}
Here, \(a_i(t)\) denotes the activation of the $i$-th muscle, \(r_i(q(t))\) is its moment arm, \(F_i(\cdot)\) represents muscle force, and \(\tau_{\mathrm{res}}(t)\) denotes reserve actuator torques. We set the activation exponent (p=2) to penalize large muscle activations and encourage distributed muscle recruitment. \(N_m\) denotes the number of muscles, which is 112 in our setting.

\subsection{Post-processing}
We apply quality filtering and temporal smoothing to improve the reliability and temporal consistency of the generated annotations. Sequences with excessive inverse-kinematics marker reconstruction errors or invalid static-optimization results are discarded. Since static optimization is performed independently at each frame, temporal smoothing is further applied to reduce high-frequency fluctuations while preserving overall activation patterns. The resulting 112-dimensional muscle activation sequences constitute the final annotations in BioHuman7M and provide supervision for BioHuman. See supplementary material for details.

\begin{figure*}[t]
  \centering
  \includegraphics[width=1.0\textwidth]{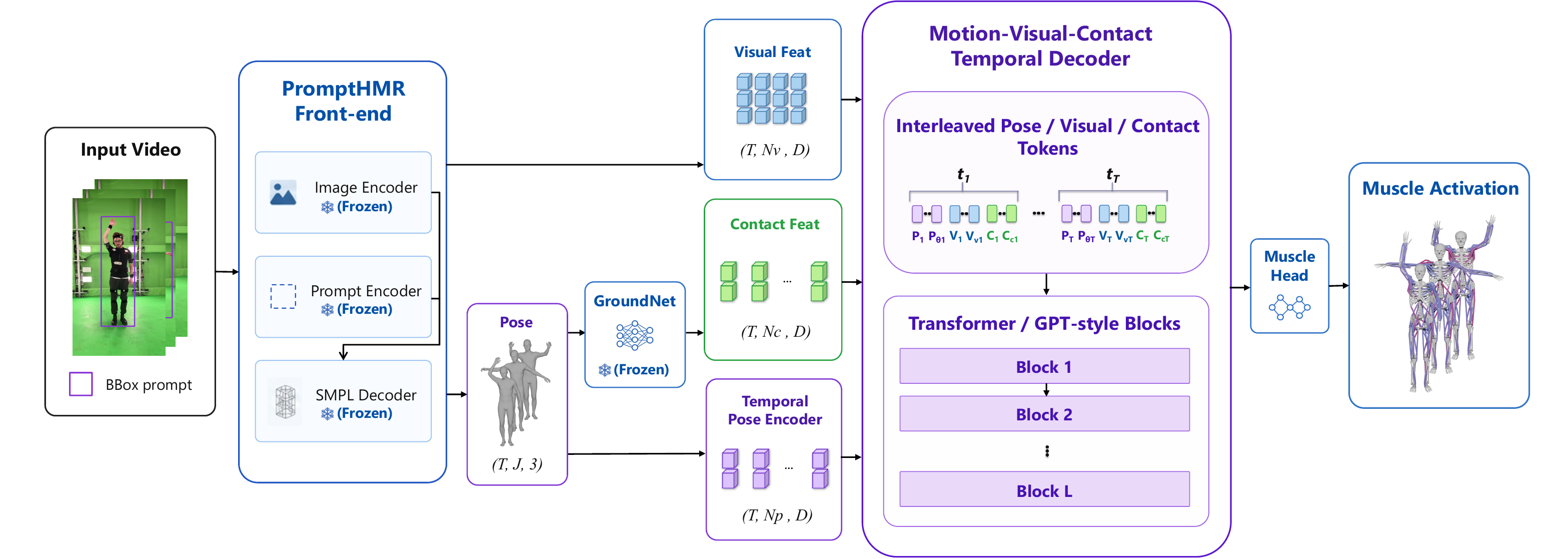}
  \caption{Pipeline of BioHuman. Given a monocular video and person bounding boxes, the HMR front-end provides pose estimates and image-conditioned visual features, while a contact encoder supplies frame-wise contact descriptors. Pose, visual, and contact tokens are processed by the temporal transformer and fused to predict full-body muscle activations with a single muscle head.}
  \label{fig:pipeline}
\end{figure*}

\section{BioHuman}\label{sec:biohuman}
We introduce BioHuman, an end-to-end video-to-muscle framework that estimates full-body muscle activations from monocular video. The framework integrates three complementary sources of evidence: human kinematics, image-level visual cues, and physical interaction information. These sources are encoded as pose, visual, and contact representations, respectively, and jointly processed within a unified framework for muscle activation estimation.

\subsection{Problem formulation}
Given a monocular video clip \(\mathbf{I}_{1:T}\) consisting of \(T\) frames and the corresponding person bounding boxes \(\mathbf{b}_{1:T}\), BioHuman aims to estimate the full-body muscle activation sequence:
\begin{equation}
  \hat{\mathbf{a}}_{1:T}
  = f_{\Theta}(\mathbf{I}_{1:T},\mathbf{b}_{1:T}),
  \qquad
  \hat{\mathbf{a}}_t \in [0,1]^{112}.
\end{equation}
where \(\hat{\mathbf{a}}_t\) denotes the predicted normalized activations of the 112 muscles in our ULBS-112 model. BioHuman is trained using the simulation-derived muscle activation annotations in BioHuman7M. OpenSim is only used to generate training supervision and is not required during inference.

\subsection{Three complementary evidence representations}\label{sec:three_tokens}
BioHuman extracts three complementary sources of evidence from monocular video: kinematic information, image-level visual features, and physical interaction cues. The first two are obtained from a pretrained HMR front-end, while the third is derived from intermediate features of the pretrained GroundNet module introduced in \cref{subsec:groundnet}.

For each frame, the HMR front-end provides an initial pose estimate and an image-conditioned feature:
\begin{equation}
  (\tilde{\mathbf{q}}_t,\mathbf{e}_t)
  = F_{\phi}\!\left(\mathcal{C}(\mathbf{I}_t,\mathbf{b}_t)\right),
\end{equation}
where $\mathcal{C}$ crops the person region. We instantiate $F_{\phi}$ with PromptHMR~\cite{wang2025prompthmr}. The entire PromptHMR front-end, including its image encoder, prompt encoder, and SMPL decoder, remains frozen during training. Its outputs serve as intermediate representations rather than explicit reconstruction targets, allowing the downstream framework to learn muscle activation estimation without modifying the pretrained HMR model.

\paragraph{Kinematic evidence.} Human motion provides informative kinematic cues about underlying muscle activations. We therefore use the estimated pose to construct an explicit representation of body articulation and temporal dynamics. Specifically, we augment the pose with first- and second-order temporal differences:
\begin{equation}
  \mathbf{r}_t =
  [\tilde{\mathbf{q}}_t,
  \Delta\tilde{\mathbf{q}}_t,
  \Delta^2\tilde{\mathbf{q}}_t],
\end{equation}

A temporal pose encoder (P) processes the resulting sequence, followed by a projection $E_p$ to obtain the pose token:
\begin{equation}
  \mathbf{z}^{p}_t = E_p\!\left(P(\mathbf{r}_{1:T})_t\right).
\end{equation}

This representation captures body configuration and local motion dynamics, providing an explicit kinematic basis for muscle activation estimation.

\paragraph{Visual evidence.} Although kinematic representations describe body articulation and motion, they may discard image-level information relevant to muscle activation estimation. We therefore retain the image-conditioned feature $\mathbf{e}_t$ from the HMR front-end as complementary visual evidence:
\begin{equation}
  \mathbf{z}^{v}_t = E_v(\mathbf{e}_t).
\end{equation}

Unlike a pose-only representation, the visual token preserves additional image-level context, including body orientation, visibility, and silhouette cues, which may complement the reconstructed kinematics.

\paragraph{Contact evidence.} Human movement also involves interactions with the surrounding environment. Contact states and external loading provide additional physical context for understanding muscle activation dynamics. We use the GroundNet module to estimate these interaction cues. Given the PromptHMR pose, global translation, and decoded heel and toe markers, GroundNet estimates bilateral contact probabilities $\boldsymbol{\kappa}_t \in \mathbb{R}^{2}$ and bilateral three-dimensional ground-reaction forces $\mathbf{f}_t \in \mathbb{R}^{6}$ in body-weight-normalized units.

Based on these estimates, we construct a 12-dimensional contact-and-loading descriptor:
\begin{equation}
  \mathbf{c}_t =
  \left[
  \boldsymbol{\kappa}_t,
  \mathbf{f}_t,
  f^R_{y,t}+f^L_{y,t},
  \frac{f^R_{y,t}}{f^R_{y,t}+f^L_{y,t}+\epsilon},
  \Delta\boldsymbol{\kappa}_t
  \right],
\end{equation}
where $f^R_{y,t}$ and $f^L_{y,t}$ denote the predicted vertical forces of the right and left feet, respectively. The descriptor captures contact states, overall loading magnitude, bilateral load distribution, and contact transitions. We then project it into the common token space:
\begin{equation}
  \mathbf{z}^{c}_t = E_c(\mathbf{c}_t).
\end{equation}

The resulting contact token provides physical interaction cues that complement the kinematic and visual representations. GroundNet remains frozen during BioHuman training. The downstream network learns to utilize the estimated contact information through muscle activation supervision.

Finally, the three token projections map their respective inputs into a common \(D\)-dimensional feature space for subsequent temporal integration.

\subsection{Unified temporal evidence integration}\label{sec:temporal_integration}
Before token interleaving, a gated projection of the visual feature is injected into the pose representation,
\begin{equation}
  \bar{\mathbf{z}}^{p}_t = \mathbf{z}^{p}_t
  + \mathbf{g}^{\mathrm{in}}_t \odot A_v(\mathbf{z}^{v}_t),
\end{equation}
where
\begin{equation}
  \mathbf{g}^{\mathrm{in}}_t = \sigma\!\left(
  M_{\mathrm{in}}[\mathbf{z}^{p}_t;A_v(\mathbf{z}^{v}_t)]
  \right).
\end{equation}
This gated residual allows image evidence to correct uncertain pose features without replacing the explicit kinematic representation. We then form the interleaved sequence
\begin{equation}
  \mathbf{Z} =
  [\bar{\mathbf{z}}^{p}_1,\mathbf{z}^{v}_1,\mathbf{z}^{c}_1,
  \ldots,
  \bar{\mathbf{z}}^{p}_T,\mathbf{z}^{v}_T,\mathbf{z}^{c}_T].
\end{equation}
Temporal positional embeddings preserve frame order, while token-type embeddings distinguish the three sources. The sequence is processed by a transformer $G_{\theta}$,
\begin{equation}
  \mathbf{O}=G_{\theta}(\mathbf{Z}).
\end{equation}
Interleaving places the three observations from each frame in a common attention neighborhood while retaining long-range access to the complete video window. Consequently, pose tokens can be interpreted using contemporaneous image and contact evidence, and ambiguous frames can draw on motion context from neighboring and temporally distant frames.

Let $\mathbf{o}^{p}_t$, $\mathbf{o}^{v}_t$, and $\mathbf{o}^{c}_t$ denote the transformer outputs associated with frame $t$. The frame-level representation uses a gated visual residual and a projected contact residual,
\begin{equation}
  \mathbf{g}^{\mathrm{out}}_t = \sigma\!\left(
  M_{\mathrm{out}}[\mathbf{o}^{p}_t;\mathbf{o}^{v}_t]
  \right),
\end{equation}
\begin{equation}
  \mathbf{h}_t = \operatorname{LN}\!\left(
  \mathbf{o}^{p}_t
  + \mathbf{g}^{\mathrm{out}}_t \odot \mathbf{o}^{v}_t
  + E^{\mathrm{out}}_c(\mathbf{o}^{c}_t)
  \right).
\end{equation}
This readout retains explicit pose, visual, and contact tokens during temporal self-attention while allowing the model to control how strongly visual and contact evidence modify the pose-centered representation used for muscle decoding.

\subsection{Muscle activation decoding and supervision}\label{subsec:loss}
The fused representation \(\mathbf h_t\) is passed to a muscle decoder \(D_a\), implemented as a multilayer perceptron (MLP), to predict normalized activations for all 112 muscles.
\begin{equation}
  \hat{\mathbf{a}}_t =
  \sigma\!\left(D_a(\mathbf{h}_t)\right),
\end{equation}
where $\sigma$ bounds the prediction to $[0,1]$.

BioHuman is trained with the muscle annotations in BioHuman7M. The objective combines activation reconstruction, temporal-difference consistency, correlation, and amplitude-sensitive terms:
\begin{equation}
  \mathcal{L} =
  \lambda_{\mathrm{rec}}\mathcal{L}_{\mathrm{rec}}
  + \lambda_{\mathrm{temp}}\mathcal{L}_{\mathrm{temp}}
  + \lambda_{\mathrm{corr}}\mathcal{L}_{\mathrm{corr}}
  + \lambda_{\mathrm{amp}}\mathcal{L}_{\mathrm{amp}}.
\end{equation}
The reconstruction term supervises activation magnitude, the temporal term preserves activation dynamics, the correlation term encourages waveform consistency, and the amplitude term emphasizes active and peak responses. 
Together, these objectives supervise both the magnitude and temporal dynamics of muscle activations. 
See supplementary material for details.
\section{Experiments}\label{sec:Experiments}  
We evaluate BioHuman through comparisons with existing methods on the simulation-based BioHuman7M benchmark, validation against real sEMG from public datasets, and ablation studies of its architectural components.

\paragraph{Implementation details.}
BioHuman is implemented in PyTorch 2.4.1 with CUDA 12.1 and PyTorch Lightning 2.4.0. Training is performed on a server equipped with eight NVIDIA A100 GPUs (80\,GB each) and two Intel Xeon Platinum 8373C CPUs, providing 72 physical cores (144 threads) in total. The main training stages use four A100 GPUs with distributed data parallelism, a batch size of 20 per GPU (an effective batch size of 80), and four data-loading workers per GPU. The final refinement is conducted on a single A100 GPU with a batch size of 32. We use 32-frame clips with a stride of 32, and resize the person-centered image input to $672 \times 672$. The model is optimized using AdamW with an initial learning rate of $5\times10^{-5}$, which is reduced to $1.2\times10^{-5}$ during the final refinement, a weight decay of $5\times10^{-4}$, and a cosine annealing schedule with a minimum learning rate of $0.1$ times the initial value. Gradients are clipped to a maximum norm of 1.0, and all experiments use bfloat16 mixed-precision training. PromptHMR, including its image encoder, prompt encoder, and SMPL decoder, remains frozen throughout training. We train for at most eight epochs with early stopping based on validation muscle-activation Pearson correlation coefficient (PCC), and use a fixed random seed of 2024. The checkpoint with the highest validation PCC is used for evaluation.

\paragraph{Datasets.} We conduct experiments on BioHuman7M and real-world sEMG datasets. BioHuman7M is used for training and simulation-based evaluation, with its training and test splits described in Section~\ref{sec:BioHuman7M}. For physiological validation, we use three public sEMG datasets: CAMS-Knee~\cite{taylor2017camsknee}, LIVE-GaitNeuroKids (LIVE)~\cite{fernandez2026lgnk}, and HUMAPS-4D~\cite{dabrowski2026humaps}. These datasets cover diverse populations and movement conditions, including gait in total knee replacement (TKR) adults, children with cerebral palsy (CP), idiopathic toe walking (ITW), and typically developing (TD) controls, as well as full-body movements in healthy adults. The recordings provide 4-16 sEMG channels depending on the dataset. See supplementary material for details.

\paragraph{Evaluation metrics.} Evaluation uses the Pearson correlation coefficient (PCC), root mean square error (RMSE), normalized root mean square error (nRMSE), mean absolute error (MAE), active MAE (aMAE), and heel-strike timing MAE (hMAE). 
We evaluate the agreement between estimated muscle activation patterns and recorded EMG signals using PCC. 
Because temporal offsets may exist between muscle activations and measured sEMG envelopes, we follow established EMG-based validation approaches~\cite{michaud2021fair} and perform bounded temporal alignment before computing PCC.
Following the convention~\cite{trinler2018estimation,cohen1992power}, we interpret $\mathrm{PCC}>0.5$ as a relatively strong positive correlation. Metrics are averaged over valid channels within each sequence and then across sequences. See supplementary material.

\paragraph{Baselines.} We compare BioHuman with two baselines: PromptHMR+MinT and OpenCap Monocular~\cite{gilon2026opencapM}. PromptHMR+MinT is a neural two-stage baseline. PromptHMR, a promptable human mesh recovery model, first estimates SMPL poses from monocular video. The resulting motion sequences are processed by MinT~\cite{schneider2024mint} to predict muscle activations. OpenCap Monocular is a video-based biomechanics pipeline that estimates 3D human kinematics from monocular video and subsequently uses OpenSim-based musculoskeletal optimization to estimate muscle activations.

\begin{table}[t]
\centering
\caption{Muscle activation estimation on BioHuman7M.}
\label{tab:comps_sim}
\scriptsize
\begin{tabular}{lcccc}
\toprule
Method & PCC$\uparrow$ & RMSE$\downarrow$ & nRMSE$\downarrow$ & aMAE$\downarrow$ \\
\midrule
PromptHMR+MinT & 0.39 & 0.070 & 0.83 & 0.117 \\
BioHuman (ours) & \textbf{0.66} & \textbf{0.064} & \textbf{0.81} & \textbf{0.076} \\
\bottomrule
\end{tabular}
\end{table}

\begin{table*}[t]
\centering
\caption{Muscle activation estimation on real sEMG datasets.}
\label{tab:semg_mint}
\scriptsize
\setlength{\tabcolsep}{2.2pt}
\resizebox{\textwidth}{!}{%
\begin{tabular}{@{}l*{12}{c}@{}}
\toprule
Method
& \multicolumn{4}{c}{CAMS-Knee}
& \multicolumn{4}{c}{LIVE}
& \multicolumn{4}{c}{HUMAPS-4D} \\
\cmidrule(lr){2-5}
\cmidrule(lr){6-9}
\cmidrule(lr){10-13}
& PCC$\uparrow$ & RMSE$\downarrow$ & nRMSE$\downarrow$ & aMAE$\downarrow$
& PCC$\uparrow$ & RMSE$\downarrow$ & nRMSE$\downarrow$ & aMAE$\downarrow$
& PCC$\uparrow$ & RMSE$\downarrow$ & nRMSE$\downarrow$ & aMAE$\downarrow$ \\
\midrule
PromptHMR+MinT
& 0.19 & 0.392 & 1.78 & 0.297
& 0.46 & 0.369 & 1.06 & 0.277
& 0.38 & 0.440 & 1.30 & 0.341 \\
OpenCap~\cite{gilon2026opencapM}
& 0.41 & 0.372 & 1.58 & 0.295 
& 0.57 & 0.350 & 0.99 & 0.299 
& 0.46 & 0.410 & 1.21 & 0.341\\
BioHuman (ours)
& \textbf{0.45} & \textbf{0.347} & \textbf{1.57} & \textbf{0.276}
& \textbf{0.68} & \textbf{0.296} & \textbf{0.85} & \textbf{0.243}
& \textbf{0.52} & \textbf{0.394} & \textbf{1.17} & \textbf{0.307} \\
\bottomrule
\end{tabular}%
}
\end{table*}

\subsection{Evaluation on simulated data}
Table~\ref{tab:comps_sim} compares BioHuman with PromptHMR+MinT on the BioHuman7M test set. Both methods use the same monocular video inputs, person bounding boxes, data split, and evaluation protocol. BioHuman consistently outperforms PromptHMR+MinT across all evaluation metrics, achieving higher correlation and lower errors.

\subsection{Evaluation on sEMG datasets}
We further validate BioHuman against real sEMG datasets, including three public datasets, covering healthy adults as well as clinical and rehabilitation populations. See supplementary material for details.

We compare BioHuman with PromptHMR+MinT and OpenCap Monocular on these datasets. \cref{tab:semg_mint} reports the corresponding quantitative results. BioHuman achieves stronger agreement with measured muscle activity, demonstrating that the learned activation patterns extend beyond fitting simulation-derived supervision. Notably, the evaluation includes diverse populations and movement conditions, including TKR adults and children with CP and ITW, providing evidence of the model's generalization to rehabilitation-relevant settings. 
Moreover, OpenCap Monocular relies on OpenSim-based musculoskeletal simulation and optimization, resulting in substantial computational overhead. It requires 36.953 s per frame, whereas BioHuman takes only 0.134 s.

\begin{figure}[t]     
\centering
\includegraphics[width=\linewidth]{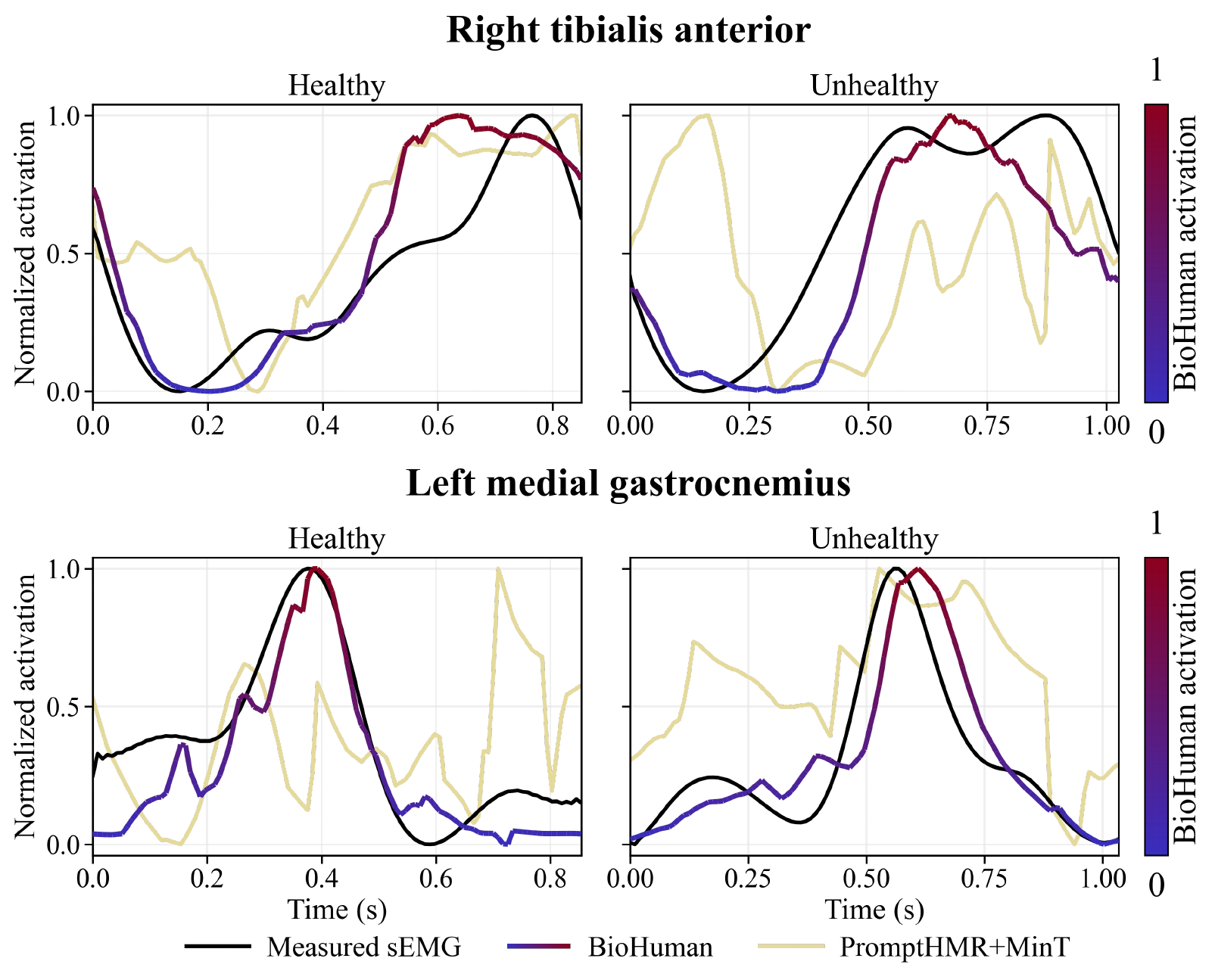}
\caption{BioHuman predictions and measured sEMG for healthy and non-healthy participants performing the same walking motion in the LIVE dataset.}
\label{fig:semg_comparison}
\end{figure}

\cref{fig:semg_comparison} presents representative activation profiles from the LIVE dataset, comparing a healthy participant and a participant with a clinical condition performing the same walking task. BioHuman captures the major activation and relaxation phases, as well as the temporal variations observed in the measured sEMG across corresponding muscles. These results provide further evidence of the physiological relevance of the predicted activation patterns and highlight the potential of video-based muscle activation estimation for gait rehabilitation and functional movement assessment.

\begin{table}[t]
\centering
\caption{Token-composition ablation on BioHuman7M and LIVE.}
\label{tab:token_composition_ablation}
\resizebox{\columnwidth}{!}{%
\begin{tabular}{@{}l*{6}{c}@{}}
\toprule
Tokens
& \multicolumn{3}{c}{BioHuman7M}
& \multicolumn{3}{c}{LIVE}\\
\cmidrule(lr){2-4}
\cmidrule(lr){5-7}
& PCC$\uparrow$ & nRMSE$\downarrow$ & hMAE$\downarrow$
& PCC$\uparrow$ & nRMSE$\downarrow$ & hMAE$\downarrow$\\
\midrule
P     & 0.56 & 0.88 & 3.208 & 0.50 & 0.88 & 3.589\\
P+V   & \textbf{0.66}&\textbf{0.81}&3.073&0.64& \textbf{0.85} & 3.589\\
P+C   &0.58& 0.87 & 2.861 & 0.61 & 0.86 & 3.344\\
P+V+C &\textbf{0.66}& \textbf{0.81}  & \textbf{2.856} &\textbf{0.68}   & \textbf{0.85}    & \textbf{3.239}\\
\bottomrule
\end{tabular}%
}
\end{table}
\subsection{Ablation}
We ablate pose (P), visual (V), and contact (C) tokens on BioHuman7M and LIVE while keeping the network architecture and training protocol fixed.
As shown in \cref{tab:token_composition_ablation}, visual tokens consistently improve activation prediction, while contact tokens provide complementary benefits, particularly in gait event timing. Combining all three token types achieves the best overall performance across both datasets, demonstrating the complementary roles of visual and contact information.

\section{Conclusions}
We presented BioHuman7M, a large-scale dataset comprising 7.46 million frames that augments existing motion resources with temporally aligned musculoskeletal annotations, including full-body muscle activations and contact-related biomechanical information. Building on this dataset, we introduced BioHuman, an end-to-end video-to-muscle framework that integrates three complementary sources of evidence: kinematic states, image-level visual observations, and physical interaction cues. By jointly modeling these representations over time, BioHuman directly estimates 112-dimensional full-body muscle activations from monocular video. Experiments on BioHuman7M demonstrate the effectiveness of the proposed framework, while validation against measured surface electromyography from clinical and rehabilitation datasets provides evidence of the physiological relevance of the estimated activation patterns. These results highlight the potential of video-based muscle activation estimation for gait rehabilitation and functional movement assessment.

\paragraph{Limitations and future work.}
BioHuman7M relies on musculoskeletal simulation to generate muscle activation annotations, and a gap remains between simulation-derived activations and actual human muscle recruitment. In particular, static optimization provides an approximate estimate of muscle recruitment under predefined optimization criteria, which may not fully capture individual differences in neuromuscular control, muscle coordination, and compensatory strategies. Moreover, surface EMG signals are correlated with muscle activation but are not directly equivalent to normalized muscle activation values. Comparing estimated activations with measured sEMG only provides indirect evidence of their physiological relevance. The current real-signal validation is further limited by the sparse channel coverage of available sEMG datasets, which prevents direct validation of all 112 predicted muscles, particularly deep muscles, as well as by the limited diversity of clinical populations and movement conditions. In addition, the current biomechanical representation is constrained by the anatomical coverage and joint degrees of freedom of the ULBS-112 model, while the contact representation primarily captures foot-ground interactions, leaving broader forms of physical interaction, such as seated support, hand contact, and object interaction, unexplored.

Future work will expand real-signal validation to larger and more diverse cohorts, movement categories, and clinical populations, while investigating subject-specific calibration, more comprehensive musculoskeletal models, and richer body-environment interaction representations. These extensions will enable more personalized assessment of muscle activation patterns and improve the applicability of BioHuman to clinical and rehabilitation settings.
\section*{LLM Usage Statement}
We used LLM to assist with the preparation of this manuscript. Specifically, it was used to refine author-written text, improve language and clarity, and polish descriptions. It also assisted with LaTeX formatting and organization. ChatGPT was not used to propose ideas, generate datasets, annotate data, execute experiments, or produce experimental results. All research decisions, implementations, experiments, results, and conclusions were conducted, verified, and approved by the authors, who take full responsibility for the submitted work.

{
    \small
    \bibliographystyle{ieeenat_fullname}
    \bibliography{main}

@String(CVPR= {IEEE Conf. Comput. Vis. Pattern Recog.})

@String(TOG= {ACM Trans. Graph.})

@String(CVPR  = {CVPR})

@String(TOG   = {ACM TOG})

@inproceedings{kanazawa2018hmr,
  title={End-to-end recovery of human shape and pose},
  author={Kanazawa, Angjoo and Black, Michael J and Jacobs, David W and Malik, Jitendra},
  booktitle={Proceedings of the IEEE/CVF Conference on Computer Vision and Pattern Recognition},
  pages={7122--7131},
  year={2018},
  address={Salt Lake City},
  publisher = "IEEE"
}

@inproceedings{zhang2021pymaf,
  title={Pymaf: 3d human pose and shape regression with pyramidal mesh alignment feedback loop},
  author={Zhang, Hongwen and Tian, Yating and Zhou, Xinchi and Ouyang, Wanli and Liu, Yebin and Wang, Limin and Sun, Zhenan},
  booktitle={Proceedings of the IEEE/CVF International Conference on Computer Vision},
  pages={11446--11456},
  year={2021},
  address={Virtual},
  publisher = "IEEE"
}

@inproceedings{zhang2024proxycap,
  title={Proxycap: Real-time monocular full-body capture in world space via human-centric proxy-to-motion learning},
  author={Zhang, Yuxiang and Zhang, Hongwen and Hu, Liangxiao and Zhang, Jiajun and Yi, Hongwei and Zhang, Shengping and Liu, Yebin},
  booktitle={Proceedings of the IEEE/CVF Conference on Computer Vision and Pattern Recognition},
  pages={1954--1964},
  year={2024},
  address={Seattle},
  publisher = "IEEE"
}

@inproceedings{kocabas2020vibe,
  title={VIBE: Video Inference for Human Body Pose and Shape Estimation},
  author={Kocabas, Muhammed and Athanasiou, Nikos and Black, Michael J.},
  booktitle={Proceedings of the IEEE/CVF Conference on Computer Vision and Pattern Recognition},
  month = {June},
  pages={5252-5262},
  year = {2020},
  address={Seattle},
  publisher = "IEEE"
}

@inproceedings{shen2024gvhmr,
  title={World-Grounded Human Motion Recovery via Gravity-View Coordinates},
  author={Shen, Zehong and Pi, Huaijin and Xia, Yan and Cen, Zhi and Peng, Sida and Hu, Zechen and Bao, Hujun and Hu, Ruizhen and Zhou, Xiaowei},
  booktitle={SIGGRAPH Asia 2024 Conference Papers},
  pages={1--11},
  address={Tokyo},
  year={2024},
  publisher={ACM}
}

@inproceedings{wang2024tram,
  title={TRAM: Global Trajectory and Motion of 3D Humans from in-the-wild Videos},
  author={Wang, Yufu and Wang, Ziyun and Liu, Lingjie and Daniilidis, Kostas},
  booktitle={Proceedings of the European Conference on Computer Vision},
  pages={467--487},
  year={2024},
  address={Milan},
  publisher={Springer}
}

@inproceedings{goel2023hmrv2,
  title={Humans in 4D: Reconstructing and tracking humans with transformers},
  author={Goel, Shubham and Pavlakos, Georgios and Rajasegaran, Jathushan and Kanazawa, Angjoo and Malik, Jitendra},
  booktitle={Proceedings of the IEEE/CVF International Conference on Computer Vision},
  pages={14783--14794},
  year={2023},
  publisher={IEEE},
  address={Paris}
}

@inproceedings{shin2024wham,
  title={Wham: Reconstructing world-grounded humans with accurate 3d motion},
  author={Shin, Soyong and Kim, Juyong and Halilaj, Eni and Black, Michael J},
  booktitle={IEEE/CVF Conference on Computer Vision and Pattern Recognition (CVPR)},
  pages={2070--2080},
  year={2024},
  address={Seattle},
  organization={IEEE}
}

@article{schneider2024mint,
  title={Muscles in time: Learning to understand human motion in-depth by simulating muscle activations},
  author={Schneider, David and Rei{\ss}, Simon and Kugler, Marco and Jaus, Alexander and Peng, Kunyu and Sutschet, Susanne and Sarfraz, M Saquib and Matthiesen, Sven and Stiefelhagen, Rainer},
  journal={Advances in Neural Information Processing Systems},
  volume={37},
  pages={67251--67281},
  year={2024}
}

@article{chen2025vid,
  title={Towards Human Inverse Dynamics from Real Images: A Dataset and Benchmark for Joint Torque Estimation},
  author={Chen, Chen and Su, Weifeng},
  journal={bioRxiv},
  pages={2025--10},
  year={2025},
  publisher={Cold Spring Harbor Laboratory}
}

@article{loper2015smpl,
  title={SMPL: A Skinned Multi-Person Linear Model},
  author={Loper, Matthew and Mahmood, Naureen and Romero, Javier and Pons-Moll, Gerard and Black, Michael J},
  journal={ACM Transactions on Graphics},
  volume={34},
  number={6},
  year={2015},
  pages={1--16},
  publisher={Association for Computing Machinery}
}

@inproceedings{keller2022osso,
  title={OSSO: Obtaining skeletal shape from outside},
  author={Keller, Marilyn and Zuffi, Silvia and Black, Michael J and Pujades, Sergi},
  booktitle={Proceedings of the IEEE/CVF conference on computer vision and pattern recognition},
  pages={20492--20501},
  year={2022},
  address = {New Orleans},
  publisher = "IEEE"
}

@article{shetty2023boss,
  title={BOSS: Bones, organs and skin shape model},
  author={Shetty, Karthik and Birkhold, Annette and Jaganathan, Srikrishna and Strobel, Norbert and Egger, Bernhard and Kowarschik, Markus and Maier, Andreas},
  journal={Computers in Biology and Medicine},
  volume={165},
  pages={107383},
  year={2023},
  publisher={Elsevier}
}

@article{keller2023skel,
  title={From skin to skeleton: Towards biomechanically accurate 3d digital humans},
  author={Keller, Marilyn and Werling, Keenon and Shin, Soyong and Delp, Scott and Pujades, Sergi and Liu, C Karen and Black, Michael J},
  journal={ACM Transactions on Graphics (TOG)},
  volume={42},
  number={6},
  pages={1--12},
  year={2023},
  publisher={ACM New York, NY, USA}
}

@article{delp2007opensim,
  title={OpenSim: open-source software to create and analyze dynamic simulations of movement},
  author={Delp, Scott L and Anderson, Frank C and Arnold, Allison S and Loan, Peter and Habib, Ayman and John, Chand T and Guendelman, Eran and Thelen, Darryl G},
  journal={IEEE transactions on biomedical engineering},
  volume={54},
  number={11},
  pages={1940--1950},
  year={2007},
  publisher={IEEE}
}

@article{bittner2022towards,
  title={Towards single camera human 3D-kinematics},
  author={Bittner, Marian and Yang, Wei-Tse and Zhang, Xucong and Seth, Ajay and Van Gemert, Jan and Van der Helm, Frans CT},
  journal={Sensors},
  volume={23},
  number={1},
  pages={341},
  year={2022},
  publisher={MDPI}
}

@article{werling2023addbiomechanics,
  title={AddBiomechanics: Automating model scaling, inverse kinematics, and inverse dynamics from human motion data through sequential optimization},
  author={Werling, Keenon and Bianco, Nicholas A and Raitor, Michael and Stingel, Jon and Hicks, Jennifer L and Collins, Steven H and Delp, Scott L and Liu, C Karen},
  journal={Plos one},
  volume={18},
  number={11},
  pages={e0295152},
  year={2023},
  publisher={Public Library of Science San Francisco, CA USA}
}

@inproceedings{lin20243d,
  title={3D kinematics estimation from video with a biomechanical model and synthetic training data},
  author={Lin, Zhi-Yi and Lyu, Bofan and Fernandez, Judith Cueto and Van Der Kruk, Eline and Seth, Ajay and Zhang, Xucong},
  booktitle={Proceedings of the IEEE/CVF Conference on Computer Vision and Pattern Recognition},
  pages={1441--1450},
  year={2024},
  address={Seattle},
  publisher = "IEEE"
}

@inproceedings{jiang2024manikin,
  title={Manikin: biomechanically accurate neural inverse kinematics for human motion estimation},
  author={Jiang, Jiaxi and Streli, Paul and Luo, Xuejing and Gebhardt, Christoph and Holz, Christian},
  booktitle={European Conference on Computer Vision},
  pages={128--146},
  year={2024},
  address={Milan},
  publisher={Springer}
}

@inproceedings{alvarado2026soma,
  title={SOMA: From Surface Observations to Muscle Anatomy},
  author={Alvarado, Eduardo and Kim, Emily and Nolte, Gerrit and Runte, Friedemann and Botsch, Mario and Habermann, Marc and Theobalt, Christian},
  booktitle={European Conference on Computer Vision},
  pages={502--520},
  year={2026},
  organization={Springer}
}

@inproceedings{gozlan2025opencapbench,
  title={OpenCapBench: A benchmark to bridge pose estimation and biomechanics},
  author={Gozlan, Yoni and Falisse, Antoine and Uhlrich, Scott and Gatti, Anthony and Black, Michael and Hicks, Jennifer and Delp, Scott and Chaudhari, Akshay},
  booktitle={2025 IEEE/CVF Winter Conference on Applications of Computer Vision (WACV)},
  pages={4056--4065},
  year={2025},
  publisher = "IEEE",
  address = {Tucson}
}

@article{pagnon2022pose2sim,
  title={Pose2Sim: An open-source Python package for multiview markerless kinematics},
  author={Pagnon, David and Domalain, Mathieu and Reveret, Lionel},
  journal={Journal of Open Source Software},
  volume={7},
  number={77},
  pages={4362},
  year={2022}
}

@inproceedings{xia2025hsmr,
  title={Reconstructing humans with a biomechanically accurate skeleton},
  author={Xia, Yan and Zhou, Xiaowei and Vouga, Etienne and Huang, Qixing and Pavlakos, Georgios},
  booktitle={Proceedings of the IEEE/CVF Conference on Computer Vision and Pattern Recognition},
  pages={5355--5365},
  year={2025},
  publisher = "IEEE",
  address = "Nashville",
}

@article{uhlrich2023opencap,
  title={OpenCap: Human movement dynamics from smartphone videos},
  author={Uhlrich, Scott D and Falisse, Antoine and Kidzi{\'n}ski, {\L}ukasz and Muccini, Julie and Ko, Michael and Chaudhari, Akshay S and Hicks, Jennifer L and Delp, Scott L},
  journal={PLoS computational biology},
  volume={19},
  number={10},
  pages={e1011462},
  year={2023},
  publisher={Public Library of Science San Francisco, CA USA}
}

@article{gilon2026opencapM,
  title={OpenCap Monocular: 3D Human Kinematics and Musculoskeletal Dynamics from a Single Smartphone Video},
  author={Gilon, Selim and Miller, Emily Y and Uhlrich, Scott D},
  journal={arXiv},
  year={2026}
}

@inproceedings{wang2025prompthmr,
  title={PromptHMR: Promptable Human Mesh Recovery},
  author={Wang, Yufu and Sun, Yu and Patel, Priyanka and Daniilidis, Kostas and Black, Michael J and Kocabas, Muhammed},
  booktitle={Proceedings of the IEEE/CVF Conference on Computer Vision and Pattern Recognition},
  pages={1148--1159},
  year={2025},
  address = "Nashville",
  publisher = "IEEE"
}

@article{ionescu2013h36m,
  title={Human3.6m: Large scale datasets and predictive methods for 3d human sensing in natural environments},
  author={Ionescu, Catalin and Papava, Dragos and Olaru, Vlad and Sminchisescu, Cristian},
  journal={IEEE transactions on pattern analysis and machine intelligence},
  volume={36},
  number={7},
  pages={1325--1339},
  year={2013},
  publisher={IEEE}
}

@inproceedings{mahmood2019amass,
  title={AMASS: Archive of motion capture as surface shapes},
  author={Mahmood, Naureen and Ghorbani, Nima and Troje, Nikolaus F and Pons-Moll, Gerard and Black, Michael J},
  booktitle={Proceedings of the IEEE/CVF international conference on computer vision},
  pages={5442--5451},
  year={2019},
  address={Seoul},
  publisher={IEEE}
}

@inproceedings{han2023groundlink,
  title={Groundlink: A dataset unifying human body movement and ground reaction dynamics},
  author={Han, Xingjian and Senderling, Ben and To, Stanley and Kumar, Deepak and Whiting, Emily and Saito, Jun},
  booktitle={SIGGRAPH Asia 2023 Conference Papers},
  pages={1--10},
  address={Sydney},
  publisher={ACM},
  year={2023}
}

@inproceedings{black2023bedlam,
  title={Bedlam: A synthetic dataset of bodies exhibiting detailed lifelike animated motion},
  author={Black, Michael J and Patel, Priyanka and Tesch, Joachim and Yang, Jinlong},
  booktitle={Proceedings of the IEEE/CVF Conference on Computer Vision and Pattern Recognition},
  pages={8726--8737},
  year={2023},
  address={Vancouver},
  publisher={IEEE}
}

@inproceedings{kaufmann2023emdb,
  title={Emdb: The electromagnetic database of global 3d human pose and shape in the wild},
  author={Kaufmann, Manuel and Song, Jie and Guo, Chen and Shen, Kaiyue and Jiang, Tianjian and Tang, Chengcheng and Z{\'a}rate, Juan Jos{\'e} and Hilliges, Otmar},
  booktitle={Proceedings of the IEEE/CVF International Conference on Computer Vision},
  pages={14632--14643},
  year={2023},
  address={Paris},
  publisher={IEEE}
}

@inproceedings{von20183dpw,
  title={Recovering accurate 3d human pose in the wild using imus and a moving camera},
  author={Von Marcard, Timo and Henschel, Roberto and Black, Michael J and Rosenhahn, Bodo and Pons-Moll, Gerard},
  booktitle={Proceedings of the European conference on computer vision},
  pages={601--617},
  year={2018},
  address={Munich},
  publisher={Springer}
}

@inproceedings{ren2025motionpro,
  title={MotionPRO: exploring the role of pressure in human mocap and beyond},
  author={Ren, Shenghao and Lu, Yi and Huang, Jiayi and Zhao, Jiayi and Zhang, He and Yu, Tao and Shen, Qiu and Cao, Xun},
  booktitle={Proceedings of the Computer Vision and Pattern Recognition Conference},
  pages={27760--27770},
  year={2025},
  address = "Nashville",
  publisher={IEEE}
}

@inproceedings{chiquier2023MiA,
  title={Muscles in action},
  author={Chiquier, Mia and Vondrick, Carl},
  booktitle={Proceedings of the IEEE/CVF International Conference on Computer Vision},
  pages={22091--22101},
  year={2023},
  address = "Vancouver",
  publisher={IEEE}
}

@article{saul2015uppermodel,
  title={Benchmarking of dynamic simulation predictions in two software platforms using an upper limb musculoskeletal model},
  author={Saul, Katherine R and Hu, Xiao and Goehler, Craig M and Vidt, Meghan E and Daly, Melissa and Velisar, Anca and Murray, Wendy M},
  journal={Computer methods in biomechanics and biomedical engineering},
  volume={18},
  number={13},
  pages={1445--1458},
  year={2015},
  publisher={Taylor \& Francis}
}

@article{van2024dataset,
  title={Dataset of running kinematics, kinetics and muscle activation at different speeds, surface gradients, cadences and with forward trunk lean},
  author={Van Hooren, Bas and Meijer, Kenneth},
  journal={Data in brief},
  volume={54},
  pages={110312},
  year={2024},
  publisher={Elsevier}
}

@inproceedings{zuo2024msh700,
  title={Self model for embodied intelligence: Modeling full-body human musculoskeletal system and locomotion control with hierarchical low-dimensional representation},
  author={Zuo, Chenhui and He, Kaibo and Shao, Jing and Sui, Yanan},
  booktitle={IEEE International Conference on Robotics and Automation (ICRA)},
  pages={13062--13069},
  year={2024},
  publisher={IEEE},
  address={Yokohama}
}

@inproceedings{dabrowski2026humaps,
  title={HUMAPS-4D: A Multimodal Dataset for HUman Motion Analysis with Physiological and Semantic informations},
  author={Dabrowski, Matthieu and Ben Jemaa, Ouala and Allaert, Benjamin},
  booktitle={Proceedings of the IEEE/CVF Conference on Computer Vision and Pattern Recognition},
  pages={21188--21197},
  year={2026},
  publisher={IEEE},
  address={Denver}
}

@article{taylor2017camsknee,
  title={A Comprehensive Assessment of the Musculoskeletal System: The {CAMS-Knee} Data Set},
  author={Taylor, William R. and Sch{\"u}tz, Pascal and Bergmann, Georg and List, Renate and Postolka, Barbara and Hitz, Marco and Dymke, J{\"o}rn and Damm, Philipp and Duda, Georg and Gerber, Hans and Schwachmeyer, Verena and Hosseini Nasab, Seyyed Hamed and Trepczynski, Adam and Kutzner, Ines},
  journal={Journal of Biomechanics},
  volume={65},
  pages={32--39},
  year={2017},
  publisher={Elsevier}
}

@article{schreiber2019multimodal,
  title={A multimodal dataset of human gait at different walking speeds established on injury-free adult participants},
  author={Schreiber, C{\'e}line and Moissenet, Florent},
  journal={Scientific data},
  volume={6},
  number={1},
  pages={111},
  year={2019},
  publisher={Nature}
}

@article{fernandez2026lgnk,
  title={LIVE-GaitNeuroKids: A multimodal and Longitudinal Gait dataset with Inertial, Video and surface Electromyography in Neuromotor conditions and typically developing children},
  author={Fern{\'a}ndez-Gonz{\'a}lez, Carmen and Saoudi, Hichem and Mazariegos-Iglesias, Celia and De la Calle, Beatriz and D{\'\i}az-Pernas, Francisco Javier and Ant{\'o}n-Rodr{\'\i}guez, M{\'\i}riam and Gonz{\'a}lez-Ortega, David and Iordanov, Daniel and Mart{\'\i}nez-Zarzuela, Mario},
  journal={Scientific Data},
  year={2026},
  publisher={Nature}
}

@article{trinler2018estimation,
  title={Estimation of muscle activation during different walking speeds with two mathematical approaches compared to surface {EMG}},
  author={Trinler, Ursula and Leboeuf, Fabien and Hollands, Kristen and Jones, Richard and Baker, Richard},
  journal={Gait \& Posture},
  volume={64},
  pages={266--273},
  year={2018},
  doi={10.1016/j.gaitpost.2018.06.115}
}

@article{cohen1992power,
  title={A Power Primer},
  author={Cohen, Jacob},
  journal={Psychological Bulletin},
  volume={112},
  number={1},
  pages={155--159},
  year={1992},
  doi={10.1037/0033-2909.112.1.155}
}

@article{michaud2021fair,
  title={A fair and {EMG}-validated comparison of recruitment criteria, musculotendon models and muscle coordination strategies, for the inverse-dynamics based optimization of muscle forces during gait},
  author={Michaud, Florian and Lamas, Mario and Lugr{\'i}s, Urbano and Cuadrado, Javier},
  journal={Journal of NeuroEngineering and Rehabilitation},
  volume={18},
  number={1},
  pages={17},
  year={2021},
  doi={10.1186/s12984-021-00806-6}
}

@article{yin2025flex,
  title={{FLEX}: A Largescale Multimodal, Multiview Dataset for Learning Structured Representations for Fitness Action Quality Assessment},
  author={Yin, Hao and Gu, Lijun and Parmar, Paritosh and Xu, Lin and Guo, Tianxiao and Liu, Xiujin and Fu, Weiwei and Zhang, Yang and Zheng, Tianyou},
  journal={arXiv},
  year={2025}
}
}
\clearpage
\setcounter{page}{1}
\maketitlesupplementary

\section{Dataset Details}
\begin{figure*}[t]
  \centering
  \includegraphics[width=0.95\textwidth]{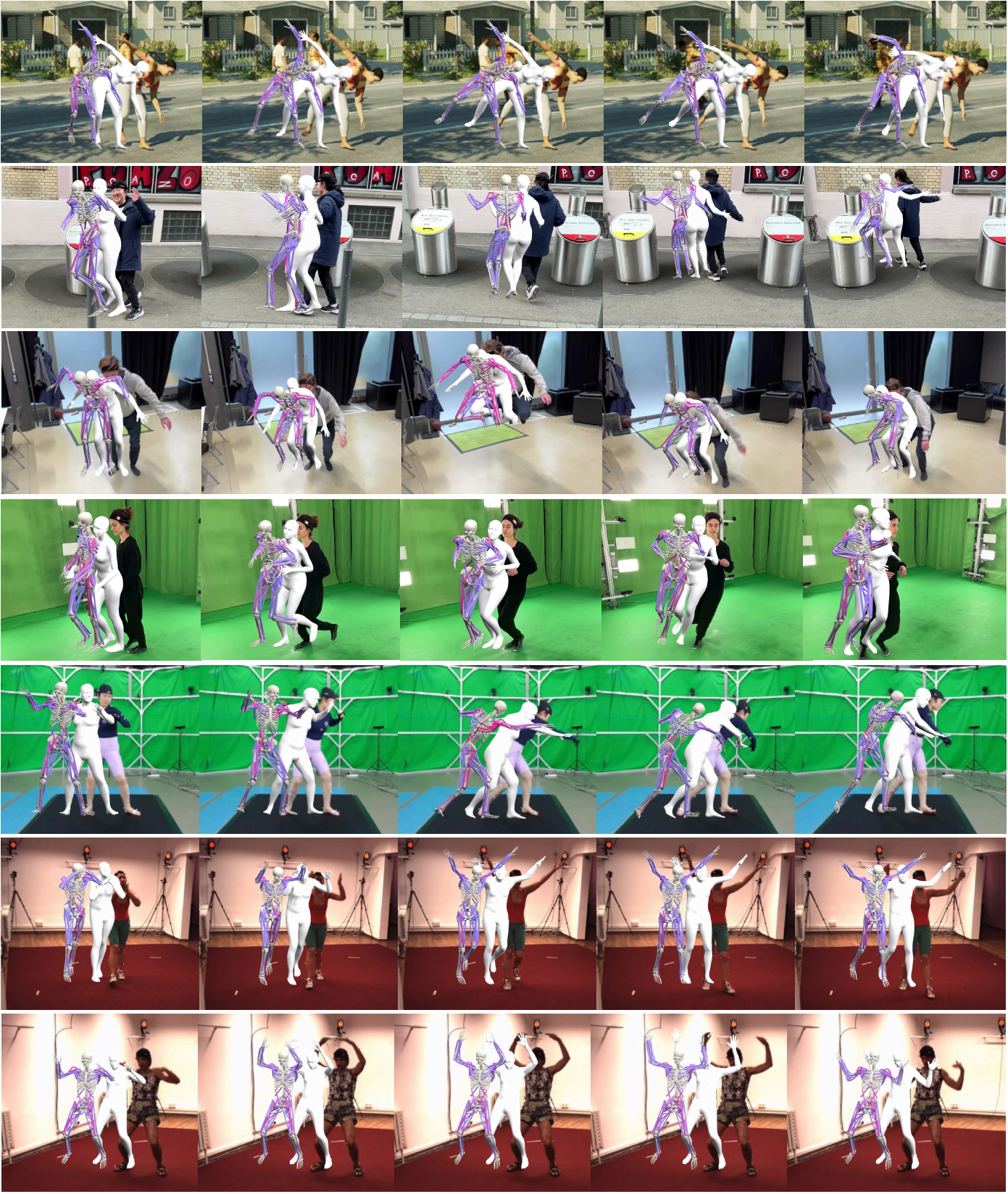}
  \caption{Example sequences from BioHuman7M.}
  \label{fig:supp_BH7M}
\end{figure*}

\paragraph{Dataset Statistics.} BioHuman7M is constructed from four source datasets: MotionPRO, EMDB, Human3.6M, and BEDLAM, as shown in Table \ref{tab:composition}. It covers a wide range of human movements, including lower limb activities such as walking, running, and jumping, as well as upper limb movements such as arm extension, waving, and punching. The video scenes cover both indoor and outdoor environments, resulting in a highly diverse dataset in terms of human actions, objects, and environmental conditions. 
Example sequences from BioHuman7M are shown in \cref{fig:supp_BH7M}.

\begin{table}[h]
\caption{Dataset composition.}
\label{tab:composition}
\centering
\begin{tabular}{lc}
  \toprule
  Dataset & Frames\\ 
  \midrule
  MotionPRO & 2.5M \\
  EMDB & 12K \\
  Human3.6M & 0.6M \\
  BEDLAM & 4.4M \\
  \midrule
  BioHuman7M (ours) & 7.46M \\
  \bottomrule
\end{tabular}
\end{table}

\paragraph{Full-body musculoskeletal model.} We develop ULBS-112, a full-body musculoskeletal model for OpenSim version 4, by adapting the official ULB model originally released for OpenSim version 3 and integrating components from MoBL-ARMS~\cite{saul2015uppermodel} and Gait2354~\cite{delp2007opensim}. ULBS-112 contains 112 muscles.

\begin{figure}[t] 
\centering \includegraphics[width=\columnwidth]{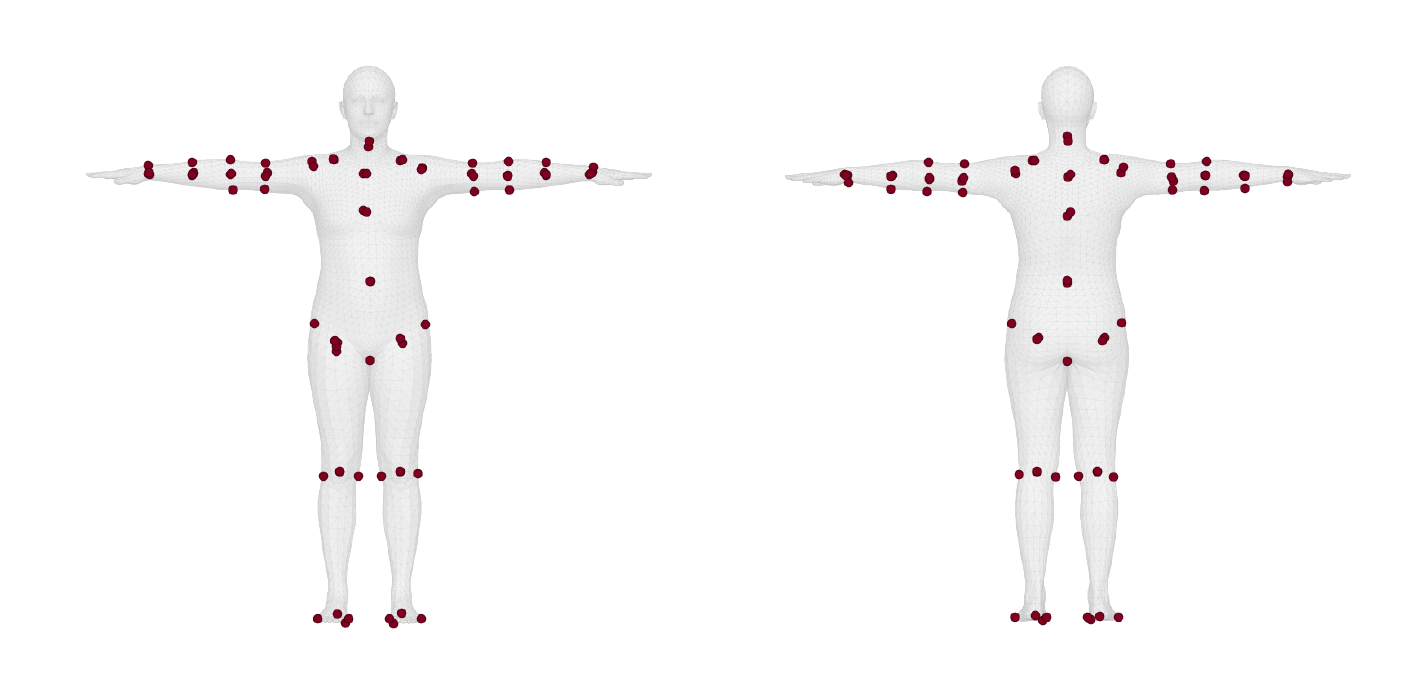} 
\caption{Locations of the 93 selected virtual markers.} \label{fig:smpl_marker_locations} 
\end{figure}

\paragraph{Marker trajectory extraction.}
We select 93 virtual markers located at bony anatomical landmarks that are relatively insensitive to soft-tissue artifacts, and export their trajectories from SMPL sequences using SMPL2AddBiomechanics. Figure~\ref{fig:smpl_marker_locations} shows the locations of the 93 selected markers.  

\paragraph{Filtering.} For the inverse kinematics results, we evaluate the marker fitting accuracy using the maximum marker error reported by OpenSim. A motion sequence is retained only if its maximum marker error is no larger than \(15\,\mathrm{cm}\). Otherwise, it is discarded. For the static optimization results, we assess the numerical validity of the solution using the constraint violation reported by OpenSim. Since static optimization solves for muscle activations under dynamic consistency constraints, a large constraint violation indicates that the estimated muscle forces and reserve actuators fail to accurately reproduce the required joint moments. We therefore retain only the sequences whose reported constraint violation is no larger than \(10^{-11}\), and discard the remaining sequences.

\paragraph{Smoothing.} Static optimization is solved independently at each time frame and does not explicitly enforce temporal continuity across adjacent frames. Consequently, frame-wise activation estimates may contain high-frequency fluctuations that are not biomechanically meaningful. To reduce such artifacts, we follow MinT~\cite{schneider2024mint} and apply a temporal smoothing procedure to all activation columns while preserving the time column. Specifically, each activation trajectory is filtered using a Savitzky-Golay filter with a window length of \(11\) frames and a polynomial order of \(2\). For \(30\,\mathrm{fps}\) data, this corresponds to a temporal window of approximately \(0.37\,\mathrm{s}\). After smoothing, activation values are clipped to the physiologically valid range \([0,1]\). The processed files are written while preserving the original OpenSim storage-file header and directory structure.

This post-processing stage ensures that the final dataset contains biomechanically plausible IK solutions, numerically valid SO results, and temporally coherent muscle activation trajectories.

\section{BioHuman Implementation Details}
\subsection{Network Details}
\paragraph{Input and supervision.}
Each input window contains RGB frames, bounding boxes, sequence identity, and frame timestamps. BioHuman estimates muscle activations $\hat{\mathbf{a}}\in[0,1]^{T\times112}$. Pose estimates from the HMR front-end are used to construct the pose tokens and, through GroundNet, the contact tokens. The HMR front-end uses the provided person boxes to crop the subject. No ground-truth pose, OpenSim state, GRF, or muscle label is used at inference time. All methods are evaluated on the same ordered windows and valid-frame masks.

\paragraph{HMR output conversion and visual feature construction.}
BioHuman instantiates the frame-level HMR front-end with PromptHMR~\cite{wang2025prompthmr}. PromptHMR estimates a 22-joint 6D rotation representation, which we convert to a 72-D SMPL axis-angle pose. A lightweight residual adapter then aligns this initial pose with the pose convention used by the unified training labels. For the image-conditioned visual feature, we use decoder-side PromptHMR features: the SMPL token and localization token are concatenated with the estimated shape and translation, and the resulting vector is projected to the BioHuman visual feature dimension before temporal fusion.

\subsection{Supervision Details} As described in \cref{subsec:loss}, the supervision comprises four terms: 
\begin{equation} 
\begin{aligned} 
\mathcal L={}&\lambda_{\mathrm{rec}}\mathcal L_{\mathrm{rec}}+\lambda_{\mathrm{temp}}\mathcal L_{\mathrm{temp}}\\ &+\lambda_{\mathrm{corr}}\mathcal L_{\mathrm{corr}}+\lambda_{\mathrm{amp}}\mathcal L_{\mathrm{amp}}. 
\end{aligned} 
\end{equation} 
where $\mathcal L_{\mathrm{rec}}$, $\mathcal L_{\mathrm{temp}}$, $\mathcal L_{\mathrm{corr}}$, and $\mathcal L_{\mathrm{amp}}$ are the reconstruction, temporal waveform, inter-muscle correlation, and amplitude-sensitive losses, and the corresponding $\lambda$ symbols are their weights.  

\paragraph{Reconstruction loss.} 
The reconstruction loss $\mathcal{L}_{\mathrm{rec}}$ penalizes discrepancies between the estimated and ground-truth SMPL poses and muscle activations:
\begin{equation} 
\mathcal L_{\mathrm{rec}}=1.0\mathsf r(\hat P,P)+1.4\mathsf r(\hat{\mathbf a},\mathbf a). 
\end{equation} 
Here, $\hat P,P\in\mathbb R^{T\times72}$ are estimated and ground-truth SMPL poses. The matrices $\hat{\mathbf a},\mathbf a\in[0,1]^{T\times112}$ are estimated and ground-truth muscle activations over $T$ frames. The stabilized RMSE is 
\begin{equation} 
\mathsf r(U,V)=\sqrt{\langle(U-V)^2\rangle+\epsilon}. 
\end{equation} 
Here, $U$ and $V$ denote the estimated and ground-truth arrays, respectively. $\langle\cdot\rangle$ averages over valid frames and labels, and $\epsilon=10^{-6}$ ensures numerical stability. A subscript restricts the averaging set.  

\paragraph{Temporal waveform loss.} 
The temporal waveform loss $\mathcal{L}_{\mathrm{temp}}$ encourages the estimated activations to reproduce the temporal dynamics and waveform patterns of the ground truth.
\begin{equation} 
\mathcal L_{\mathrm{temp}}=\mathcal T_{\mathrm{muscle}}+\mathcal C_{\mathrm{wave}}. \end{equation} 

The muscle-dynamics term $\mathcal{T}_{\mathrm{muscle}}$ matches local activation changes and centered waveforms while regularizing temporal smoothness: 
\begin{equation} 
\begin{aligned} 
\mathcal T_{\mathrm{muscle}}={}&2.0\langle|\Delta\hat{\mathbf a}-\Delta\mathbf a|\rangle +1.0\mathsf r(\hat{\mathbf a}',\mathbf a')\\ &+1.0\operatorname{mean}_c(w_cq_c) +0.5\frac{\sum_{t,c}s_c(t)e'_{t,c}}{\sum_{t,c}s_c(t)}\\ &+0.5\operatorname{mean}_{c\in\mathcal D_E}|\hat E_c-E_c| +0.05\langle|\Delta^2\hat{\mathbf a}|\rangle. \end{aligned} 
\end{equation} 
Here, $t$ indexes frames and $c\in\{1,\ldots,112\}$ indexes muscle channels. The vectors $\mathbf a_c,\hat{\mathbf a}_c$ are channel-wise time series, with scalar entries $a_{t,c},\hat a_{t,c}$. For a time-indexed quantity $z_t$, $\Delta z_t=z_t-z_{t-1}$ and $\Delta^2z_t=z_t-2z_{t-1}+z_{t-2}$. All contributing frames must be valid. Primes indicate window-centered signals $\mathbf a'_c=\mathbf a_c-\bar{\mathbf a}_c$ and $\hat{\mathbf a}'_c=\hat{\mathbf a}_c-\bar{\hat{\mathbf a}}_c$. Bars and $\sigma$ denote the relevant mean and standard deviation. The pointwise error is $e'_{t,c}=|\hat a'_{t,c}-a'_{t,c}|$, and $q_c=\mathsf r(\hat{\mathbf a}'_c,\mathbf a'_c)$ is its channel RMSE. Dynamic-channel and frame-salience weights are 
\begin{equation} 
\begin{aligned} 
u_c&=\operatorname{clip}(\sigma(\mathbf a_c)/0.01,1,5),\\ w_c&=u_c/\left(\frac1{112}\sum_j u_j\right),\\ s_c(t)&=1+\frac{2|a'_{t,c}|}{\sigma(\mathbf a_c)+0.01}. 
\end{aligned} 
\end{equation} 
Here, $j$ indexes muscle channels and $\operatorname{clip}(x,l,u)=\min\{\max(x,l),u\}$ clamps $x$ to $[l,u]$. The salience weights are treated as constants during backpropagation. The difference energies are $E_c=\sqrt{\langle(\Delta\mathbf a_c)^2\rangle+\epsilon}$ and $\hat E_c=\sqrt{\langle(\Delta\hat{\mathbf a}_c)^2\rangle+\epsilon}$. The set $\mathcal D_E$ contains channels with sufficient ground-truth difference energy. Centering, channel weighting, salience weighting, energy comparison, and smoothness are computed separately within each window and channel, using valid entries. 

The waveform-correlation term $\mathcal{C}_{\mathrm{wave}}$ encourages temporal agreement between the estimated and ground-truth activation signals: 
\begin{equation} 
\begin{aligned} 
\mathcal C_{\mathrm{wave}}={}&20.0\left(1-\frac1{112}\sum_c\rho_c\right)\\ &+7.0\operatorname{mean}_{c:\sigma(\mathbf a_c)>0.008}(1-\rho_c^{\mathrm{win}}). 
\end{aligned} 
\end{equation} 
Here, $\rho_c$ is the PCC between the estimated and ground-truth muscle activation signals for channel $c$, and $\rho_c^{\mathrm{win}}$ is computed within a window. The second contribution uses the window-level ground-truth standard-deviation threshold of 0.008. PCC is 
\begin{equation} 
\rho_c=\frac{\langle(\hat{\mathbf a}_c-\bar{\hat{\mathbf a}}_c)(\mathbf a_c-\bar{\mathbf a}_c)\rangle}{\sigma(\hat{\mathbf a}_c)\sigma(\mathbf a_c)+\epsilon}. 
\end{equation} 
The means, standard deviations, and covariance use the same valid time points.  

\paragraph{Inter-muscle correlation.} 
The inter-muscle correlation loss $\mathcal{L}_{\mathrm{corr}}$ preserves muscle coordination patterns by matching temporal correlations between muscle channels:
\begin{equation} 
\mathcal L_{\mathrm{corr}}=1.8\langle|\hat R^{\mathrm{corr}}_{ij}-R^{\mathrm{corr}}_{ij}|\rangle_{i<j}. 
\end{equation} 
Here, $\hat R^{\mathrm{corr}},R^{\mathrm{corr}}$ are estimated and ground-truth temporal correlation matrices across muscle channels. Indices $i,j$ identify distinct channels. Only channels with ground-truth standard deviation at least 0.01 and pairs $i<j$ are used.  

\paragraph{Amplitude-sensitive supervision.} 
The amplitude-sensitive loss $\mathcal{L}_{\mathrm{amp}}$ emphasizes activation accuracy at informative time points and agreement in activation statistics:
\begin{equation} 
\mathcal L_{\mathrm{amp}}=\mathcal A_{\mathrm{response}}+\mathcal A_{\mathrm{stats}}. \end{equation}

The activation-response term $\mathcal{A}_{\mathrm{response}}$ supervises active, peak, and low-activation responses, together with activity gating and muscle recruitment patterns: 
\begin{equation} 
\begin{aligned} 
\mathcal A_{\mathrm{response}}={}&1.2D(\Omega_{\mathrm{active}})+1.2D(\Omega_{\mathrm{peak}})\\ &+1.2\operatorname{mean}_{w,c}D_{wc}(\Omega^{\mathrm{high}}_{wc})\\ &+2.5\operatorname{mean}_{w,c}D_{wc}(\Omega^{\mathrm{low}}_{wc})\\ &+0.25\langle\hat a_{t,c}^2\rangle_{a_{t,c}\leq0.05}\\ &+0.5B_p(G,A)+0.25B(G,A_{\mathrm{soft}})\\ &+0.25\left[B_{\mathcal D_\phi}(G,A_{\mathrm{phase}})+P_{\mathrm{margin}}\right]\\ &+1.5\langle|\hat r_c(t)-r_c(t)|\rangle_{\mathrm{leg\text{-}active}}\\ &+1.0\langle|\hat\ell(t)-\ell(t)|\rangle\\ &+0.6\langle|\hat a_{t,i}\hat a_{t,j}-a_{t,i}a_{t,j}|\rangle_{(i,j)\in\mathcal P}. 
\end{aligned} 
\end{equation} 
Here, $D(S)=\langle|\hat a_{t,c}-a_{t,c}|\rangle_S$ is MAE on a selected valid set $S$, and $D_{wc}$ restricts it to window $w$ and channel $c$. The set $\Omega_{\mathrm{active}}$ contains entries with $a_{t,c}>0.05$, and $\Omega_{\mathrm{peak}}$ contains the highest 10\% of all valid ground-truth entries. The sets $\Omega^{\mathrm{high}}_{wc},\Omega^{\mathrm{low}}_{wc}$ contain the highest and lowest 10\% of valid ground-truth frames in each window and channel. Each selected set is normalized independently. The training threshold 0.05 differs from the evaluation aMAE threshold 0.10. The functions $B$ and $B_p$ denote BCEWithLogits without and with a positive-class weight, respectively. The logit array is $G$. The positive-class weight $p$ balances valid positive and negative entries in the batch and is clipped to $[0.25,4.0]$. The three gate targets are 
\begin{equation} 
\begin{aligned} 
A&=\mathbf1[\mathbf a>0.05],\\ A_{\mathrm{soft}}&=\operatorname{clip}(\mathbf a/0.3,0,1),\\ A_{\mathrm{phase}}&=\left[\operatorname{clip}\!\left(\frac{\mathbf a-Q_{0.1}(\mathbf a)}{Q_{0.9}(\mathbf a)-Q_{0.1}(\mathbf a)+\epsilon},0,1\right)\right]^{2.5}. 
\end{aligned} 
\end{equation} 
Here, $\mathbf1[\cdot]$ is the indicator function, equal to 1 when its condition holds and 0 otherwise. Comparisons and clipping act element-wise. The operator $Q_q$ is the $q$-quantile over valid times within each window and channel. The constant $\epsilon$ prevents division by zero. The set $\mathcal D_\phi$ contains window-channel pairs with $Q_{0.9}-Q_{0.1}\geq0.02$. The term $P_{\mathrm{margin}}$ applies the existing soft low-/high-phase boundary constraints with margin 0.1 on the same eligible pairs. For each leg's set $\mathcal C_\ell$ of 24 lower-limb muscles, $r_c(t)=a_{t,c}/(\sum_{j\in\mathcal C_\ell}a_{t,j}+\epsilon)$ is its recruitment proportion, and $\ell(t)=\frac1{24}\sum_{c\in\mathcal C_\ell}a_{t,c}$ is leg-average activation. Hats denote the predicted counterparts. The recruitment-proportion average uses frames with leg-total ground-truth activation at least 0.1. The set $\mathcal P$ contains the predefined key antagonist pairs. Their product error matches observed coactivation. 

The activation-statistics term $\mathcal{A}_{\mathrm{stats}}$ matches activation levels and dynamic ranges while discouraging overly flat estimates: 
\begin{equation} 
\begin{aligned} 
\mathcal A_{\mathrm{stats}}={}&2.5\operatorname{mean}_c|\hat v_c-v_c|\\ &+1.2\operatorname{mean}_{w,c}|\hat v_{wc}-v_{wc}|\\ &+1.2\operatorname{mean}_{w,c}|\hat R_{wc}-R_{wc}|\\ &+0.1\operatorname{mean}_{w,c}|\hat\mu_{wc}-\mu_{wc}|\\ &+0.5\operatorname{mean}_{w,c}|\hat b_{wc}-b_{wc}|\\ &+1.0\operatorname{mean}_{(w,c)\in\mathcal D_s}\left|\log\frac{\hat v_{wc}+0.001}{v_{wc}+0.001}\right|\\ &+1.0\operatorname{mean}_{(w,c)\in\mathcal D_s}\left|\log\frac{\hat R_{wc}+0.001}{R_{wc}+0.001}\right|\\ &+0.8\operatorname{mean}_{(w,c)\in\mathcal D_s}[0.9v_{wc}-\hat v_{wc}]_+. 
\end{aligned} 
\end{equation} 
Here, $v_c=\sigma(\mathbf a_c)$ is channel-level standard deviation. The vector $\mathbf a_{wc}$ is the valid ground-truth time series within window $w$ and channel $c$. Within that window and channel, $v_{wc}=\sigma(\mathbf a_{wc})$, $R_{wc}=Q_{0.95}(\mathbf a_{wc})-Q_{0.05}(\mathbf a_{wc})$, $\mu_{wc}=\overline{\mathbf a}_{wc}$, and $b_{wc}=Q_{0.05}(\mathbf a_{wc})$ denote standard deviation, quantile range, mean, and baseline. Hats denote the corresponding predicted statistics. The set $\mathcal D_s$ contains pairs with $v_{wc}>0.015$, $[x]_+=\max(x,0)$, and 0.001 stabilizes the logarithms. The last term penalizes only predicted standard deviations below 90\% of the reference. All statistics and averages use valid entries and eligible channel sets.  

\paragraph{Mixed-dataset training.}
For mixed-dataset training, we aggregate losses using dataset-specific weights: 
\begin{equation} 
\mathcal{L}_{\mathrm{train}}= \frac{\sum_d n_dw_d\mathcal{L}_d}{\sum_d n_d}, 
\end{equation} 
where $\mathcal{L}_d$ denotes the total loss defined above evaluated on samples from dataset $d$, and $n_d$ is the number of samples from that dataset. The dataset weights are $w_{\mathrm{MotionPRO}}=1.0$, $w_{\mathrm{BEDLAM}}=1.25$, $w_{\mathrm{Human3.6M}}=1.6$, and $w_{\mathrm{EMDB}}=1.0$. The retained pose reconstruction coefficient is 1.0. Evaluation metrics are computed independently as described in \cref{supp:expde} (Metrics). 

Shorter sequences are padded with additional frames to match the common window length for batched processing, and these frames do not represent actual observations. All muscle losses are evaluated only on valid labels. The muscle activation validity mask $\mathbf{m}^{a}$ excludes padded frames and frames whose OpenSim-derived muscle activation labels do not meet our quality criteria. Temporal-difference terms are computed only for adjacent frame pairs for which both frames are valid. The temporal transformer uses a padding mask derived from the window lengths, whereas the muscle-validity mask is applied to the losses and evaluation metrics. For the muscle PCC loss, PCC is computed separately for each valid muscle channel within each sequence using the valid time points selected by $\mathbf{m}^{a}$. The correlations are first averaged over valid channels within each sequence and then over sequences in the current mini-batch. This prevents padded frames and invalid muscle activation labels from contributing to the training objective.

\begin{table*}[t]
\centering
\caption{Evaluation of the original pretrained PromptHMR+MinT pipeline.}
\label{tab:original_mint}
\scriptsize
\setlength{\tabcolsep}{2.2pt}
\resizebox{\textwidth}{!}{%
\begin{tabular}{@{}l*{16}{c}@{}}
\toprule
Method
& \multicolumn{4}{c}{BioHuman7M}
& \multicolumn{4}{c}{CAMS-Knee}
& \multicolumn{4}{c}{LIVE}
& \multicolumn{4}{c}{HUMAPS-4D} \\
\cmidrule(lr){2-5}
\cmidrule(lr){6-9}
\cmidrule(lr){10-13}
\cmidrule(lr){14-17}
& PCC$\uparrow$ & RMSE$\downarrow$ & nRMSE$\downarrow$ & aMAE$\downarrow$
& PCC$\uparrow$ & RMSE$\downarrow$ & nRMSE$\downarrow$ & aMAE$\downarrow$
& PCC$\uparrow$ & RMSE$\downarrow$ & nRMSE$\downarrow$ & aMAE$\downarrow$
& PCC$\uparrow$ & RMSE$\downarrow$ & nRMSE$\downarrow$ & aMAE$\downarrow$ \\
\midrule
PromptHMR+MinT(original)
& 0.17 & 0.095 & 1.02 & 0.148
& 0.18 & 0.483 & 2.19 & 0.378
& 0.27 & 0.500 & 1.46 & 0.403
& 0.23 & 0.497 & 1.42 & 0.464 \\
PromptHMR+MinT(finetune)
& 0.39 & 0.070 & 0.83 & 0.117
& 0.19 & 0.392 & 1.78 & 0.297
& 0.46 & 0.369 & 1.06 & 0.277
& 0.38 & 0.440 & 1.30 & 0.341 \\
BioHuman (ours)
& \textbf{0.66} & \textbf{0.064} & \textbf{0.81} & \textbf{0.076}
& \textbf{0.45} & \textbf{0.347} & \textbf{1.57} & \textbf{0.276}
& \textbf{0.68} & \textbf{0.296} & \textbf{0.85} & \textbf{0.243}
& \textbf{0.52} & \textbf{0.394} & \textbf{1.17} & \textbf{0.307} \\
\bottomrule
\end{tabular}%
}
\end{table*}

\section{Experimental Details}\label{supp:expde}

\paragraph{Train/test split.}
BioHuman7M is split at the sequence level to avoid frame leakage between training and testing. MotionPRO, BEDLAM, Human3.6M, and EMDB follow the same fixed approximately 8:2 sequence-level split used in our running training configuration. During training, fresh random 10\% and 8\% subsets of the MotionPRO and BEDLAM training windows, respectively, are sampled in each epoch, while Human3.6M and EMDB use their full training-window sets. The exported test protocol contains temporal windows, each with up to 32 frames, official or aligned person boxes, SMPL pose labels, and 112-D muscle activation labels.

\paragraph{Real sEMG Datasets.}
As summarized in \cref{tab:semg_datasets}, CAMS-Knee provides 16-channel lower-limb sEMG recordings from adults with total knee replacement (TKR). We use only the sequences publicly available on the dataset website, as access to the remaining data requires application approval. LIVE contains four-channel lower-limb sEMG recordings from children with cerebral palsy (CP), idiopathic toe walking (ITW), and typical development (TD). HUMAPS-4D provides 16-channel sEMG recordings spanning upper and lower body regions from healthy adults.
\begin{table}[t]
    \centering
    \caption{sEMG datasets used in our evaluation.}
    \label{tab:semg_datasets}
    \scriptsize
    \begin{tabular*}{\columnwidth}{@{\extracolsep{\fill}}lccccc@{}}
        \toprule
        Dataset & Region & Ch. & Cohort \\
        \midrule
        CAMS-Knee & Lower & 16 & TKR adults \\
        LIVE & Lower & 4 & CP/ITW/TD children \\
        HUMAPS-4D & Full & 16 & Healthy adults \\
        \bottomrule
    \end{tabular*}
\end{table}  

\paragraph{Metrics.}
We evaluate muscle activations using the Pearson correlation coefficient (PCC), root mean square error (RMSE), normalized RMSE (nRMSE), active mean absolute error (aMAE) at a threshold of 0.10, and heel-strike timing mean absolute error (hMAE).

Muscle quality is measured over the 112-D BioHuman muscle activation space. Let $\Omega$ denote the valid set of time-muscle entries selected by the evaluation mask.

PCC is computed for each valid muscle channel within each sequence,
then averaged over valid channels within each sequence and across sequences. It measures the temporal waveform agreement between predicted and reference activation signals.

The muscle RMSE measures absolute activation accuracy and emphasizes large errors. It is defined as
\begin{equation}
  \mathrm{RMSE} =
  \sqrt{\frac{1}{|\Omega|}\sum_{(t,c)\in\Omega}
  (\hat{a}_{t,c}-a_{t,c})^2}.
\end{equation}

We also report normalized RMSE,
\begin{equation}
  \mathrm{nRMSE}=\frac{\mathrm{RMSE}}{\mathrm{Std}(a_{t,c}:(t,c)\in\Omega)} ,
\end{equation}
which normalizes the magnitude error by the standard deviation of the
ground-truth activations.

We report active MAE (aMAE) at a threshold of 0.10, defined as the mean absolute error over valid entries whose ground-truth activation exceeds 0.10. This active-entry metric emphasizes nontrivial muscle firing rather than the many near-zero background entries.

We also report the Heel-strike Timing MAE (hMAE), which evaluates the temporal accuracy of predicted muscle-activation responses at foot contact. Heel strike is identified when the ipsilateral vertical ground-reaction force crosses the 20-N threshold. Within a $\pm5$-frame window around each event, hMAE measures the absolute temporal offset between the maximum activation-rise rates of the predicted and ground-truth loading muscles, averaged across all valid events. Lower values indicate more accurately timed muscular responses to heel strike.

\paragraph{Baseline details.}
We compare BioHuman with two representative approaches: a neural two-stage baseline and a biomechanics-based pipeline. For the neural baseline, we construct PromptHMR+MinT by combining PromptHMR for monocular human pose and motion reconstruction with a MinT-style temporal transformer for muscle activation prediction.  This baseline represents a conventional sequential pipeline that estimates muscle activations solely from reconstructed motion. We also compare against OpenCap Monocular~\cite{gilon2026opencapM}, which estimates 3D human kinematics from a single smartphone video and subsequently employs an OpenSim-based musculoskeletal pipeline to recover muscle activations.

\begin{figure*}[t]
  \centering
  \includegraphics[width=0.95\textwidth]{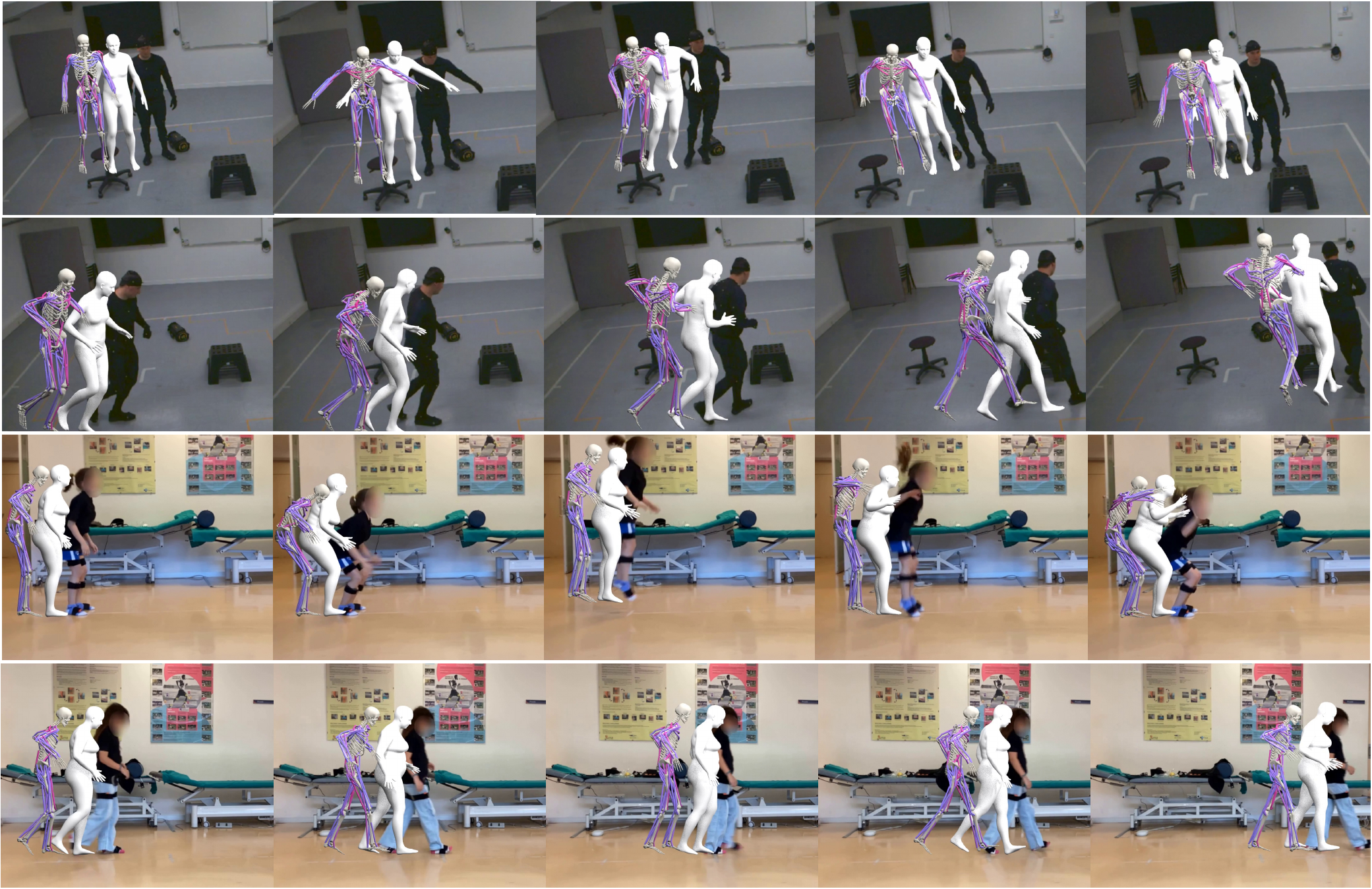}
  \caption{Example qualitative results of BioHuman on test sequences.}
  \label{fig:supp_result}
\end{figure*}

\section{Additional experimental results}

\paragraph{BioSim validation.}
Since BioSim is a simulation pipeline applied to public datasets, we additionally evaluate the reliability of its outputs using the 138\_HealthyPiG subset of the dataset \emph{A full-body motion capture gait dataset of 138 able-bodied adults across the life span and 50 stroke survivors}. In the absence of a suitable public dataset providing both SMPL parameters and sEMG recordings, we use 138\_HealthyPiG, which provides marker trajectories and sEMG. We fit SMPL to its marker trajectories and use the fitted SMPL sequences as input to BioSim to obtain muscle activations. With appropriate temporal alignment~\cite{michaud2021fair}, the resulting activations achieve a PCC of 0.84 with the measured sEMG reference signals, providing evidence for the reliability of BioSim. The SMPL fits used in this validation are of lower quality than those provided by the source datasets used to construct BioHuman7M. More accurate SMPL inputs may therefore improve agreement, although this possibility is not directly established by the present experiment.

\paragraph{Evaluation with the original MinT model.} We evaluate the original pretrained PromptHMR+MinT pipeline and its fine-tuned variant on BioHuman7M, CAMS-Knee, HUMAPS-4D, and LIVE. For a fair comparison, all methods use the same frozen PromptHMR front-end with its original pretrained weights. The pretrained variant uses the original MinT weights without further training, whereas the fine-tuned variant trains MinT on BioHuman7M. BioHuman is evaluated using the same frozen PromptHMR front-end. Table~\ref{tab:original_mint} reports the corresponding PCC, RMSE, nRMSE, and aMAE results.

\paragraph{Qualitative Results.}~\cref{fig:supp_result} presents qualitative examples of BioHuman predictions on test sequences. Given monocular video inputs, BioHuman estimates temporally varying activations across the full body, capturing muscle activation patterns throughout different movements. These visualizations illustrate the model's ability to translate visual motion observations into biomechanical representations of human movement.

\section{Reproducibility}
BioHuman7M is constructed from existing publicly available motion datasets, including Human3.6M, BEDLAM, EMDB, and MotionPRO. The original data remain subject to the respective licenses and terms of use established by their providers. We do not claim ownership of these source datasets or redistribute their original content beyond the permissions granted by the respective licenses.

Upon publication, we will publicly release all the pipeline, model weights, training and evaluation code, and derived musculoskeletal annotations and metadata to the extent permitted by the licenses of the source datasets. For datasets with redistribution restrictions, users will be directed to obtain the original data from their respective providers. We will also provide the necessary processing scripts and instructions to facilitate reproduction.

\end{document}